\documentclass[journal]{IEEEtran}
\usepackage[english]{babel}
\usepackage{times}
\usepackage{graphicx}
\usepackage{amsmath}
\usepackage{amssymb}
\usepackage{epsfig}
\usepackage{multirow}
\usepackage{float}
\usepackage{indentfirst}
\usepackage{marvosym}
\usepackage{authblk}
\usepackage{fix-cm}
\usepackage{cite}
\usepackage{caption}
\usepackage[backref]{hyperref}
\usepackage{setspace}
\usepackage{subcaption}
\usepackage{makecell}
\usepackage{booktabs}
\usepackage{color}
\usepackage{algorithm}
\usepackage{algpseudocode}
\usepackage{amsmath}
\usepackage{colortbl} 
\usepackage{xcolor}
\usepackage{array}
\usepackage{soul} 
\usepackage{picins}
\usepackage[T1]{fontenc} 
\definecolor{lightyellow}{RGB}{135,208,235}
\definecolor{yellow}{RGB}{64,224,225}
\definecolor{orange}{RGB}{100,160,225}
\usepackage{tabularx}
\usepackage{graphicx}
\usepackage{amssymb}
\usepackage{pifont}
\usepackage{wasysym}
\usepackage{utfsym}
\usepackage{fontawesome}
\usepackage{cancel}
\usepackage{colortbl}
\definecolor{colorh}{RGB}{105, 255, 180}
\definecolor{colorm}{RGB}{145, 255, 255}
\definecolor{colorl}{RGB}{225, 255, 255}
\newcommand{\colorh}[1]{\colorbox{colorh}{{#1}}}
\newcommand{\colorm}[1]{\colorbox{colorm}{{#1}}}
\newcommand{\colorl}[1]{\colorbox{colorl}{{#1}}}

\begin{document}
\makeatletter
\renewcommand{\maketag@@@}[1]{\hbox{\m@th\normalsize\normalfont#1}}%
\makeatother
\newcommand{\myfont}{\fontsize{7.0pt}{\baselineskip}\selectfont}
\newcommand{\testfont}{\fontsize{7.0pt}{\baselineskip}\selectfont}
\newcommand{\smallfont}{\fontsize{7.0pt}{\baselineskip}\selectfont}
\newcommand{\tablefont}{\fontsize{8.4pt}{\baselineskip}\selectfont}
\newcommand{\tablefontsmall}{\fontsize{6.0pt}{\baselineskip}\selectfont}
\newcommand{\smallfontmy}{\fontsize{7.5pt}{\baselineskip}\selectfont}
\captionsetup[figure]{labelformat={default},labelsep=period,name={Figure}}
\title{UniQuadric: A SLAM Backend for Unknown Rigid Object 3D Tracking and Light-Weight Modeling}
\author{
    Linghao Yang\textsuperscript{\rm 1} \quad ~
    Yanmin Wu\textsuperscript{\rm 2} \quad ~
    Yu Deng\textsuperscript{\rm 1} \quad ~
    Rui Tian\textsuperscript{\rm 3} \quad ~
    Xinggang Hu\textsuperscript{\rm 4} \quad ~
    Tiefeng Ma\textsuperscript{\rm 1}$^*$
    \vspace{-5pt}
    \\
    \textsuperscript{\rm 1} Xi'an Precision Machinery Research Institute Kunming Branch, China \\
    \textsuperscript{\rm 2} Shenzhen Graduate School, Peking University, China \quad
    \textsuperscript{\rm 3} Northeastern University, China \\
    \textsuperscript{\rm 4} Dalian University of Technology, China \\
    {\tt\small linghaoyangneu@163.com}
    \vspace{-2em}
}
\maketitle
\renewcommand{\thefootnote}{}
\footnotetext{$^*$\textit{Corresponding author}. This work was supported by COIF (No. JJ-2020-705-02)}

\begin{spacing}{1.2}
\begin{abstract}
Tracking and modeling unknown rigid objects in the environment play a crucial role in autonomous unmanned systems and virtual-real interactive applications. However, many existing Simultaneous Localization, Mapping and Moving Object Tracking (SLAMMOT) methods focus solely on estimating specific object poses and lack estimation of object scales and are unable to effectively track unknown objects.
In this paper, we propose a novel SLAM backend that unifies ego-motion tracking, rigid object motion tracking, and modeling within a joint optimization framework.
In the perception part, we designed a pixel-level asynchronous object tracker (AOT) based on the Segment Anything Model (SAM) and DeAOT, enabling the tracker to effectively track target unknown objects  guided by various predefined tasks and prompts.
In the modeling part, we present a novel object-centric quadric parameterization to unify both static and dynamic object initialization and optimization.
Subsequently, in the part of object state estimation, we propose a tightly coupled optimization model for object pose and scale estimation, incorporating hybrids constraints into a novel dual sliding window optimization framework for joint estimation.
To our knowledge, we are the first to tightly couple object pose tracking with light-weight modeling of dynamic and static objects using quadric. We conduct qualitative and quantitative experiments on simulation datasets and real-world datasets, demonstrating the state-of-the-art robustness and accuracy in motion estimation and modeling. Our system showcases the potential application of object perception in complex dynamic scenes.
\end{abstract}
\end{spacing}
\def\IEEEkeywordsname{Keywords}
%
\IEEEpeerreviewmaketitle
\linespread{1.3}
\vspace{-1em}
\section{\bfseries Introduction}



\IEEEPARstart{A}{ccurate} and robust perception of ego-motion and the motion and scale of other objects in the surrounding environment is a key technology for autonomous unmanned system navigation, obstacle avoidance, and virtual/augmented reality interactions.

\begin{figure}[t]
\setlength{\abovecaptionskip}{0.cm}
\setlength{\belowcaptionskip}{-1.0cm}
\setcounter{figure}{0}
	\centering
	\includegraphics[width=1.0\linewidth]{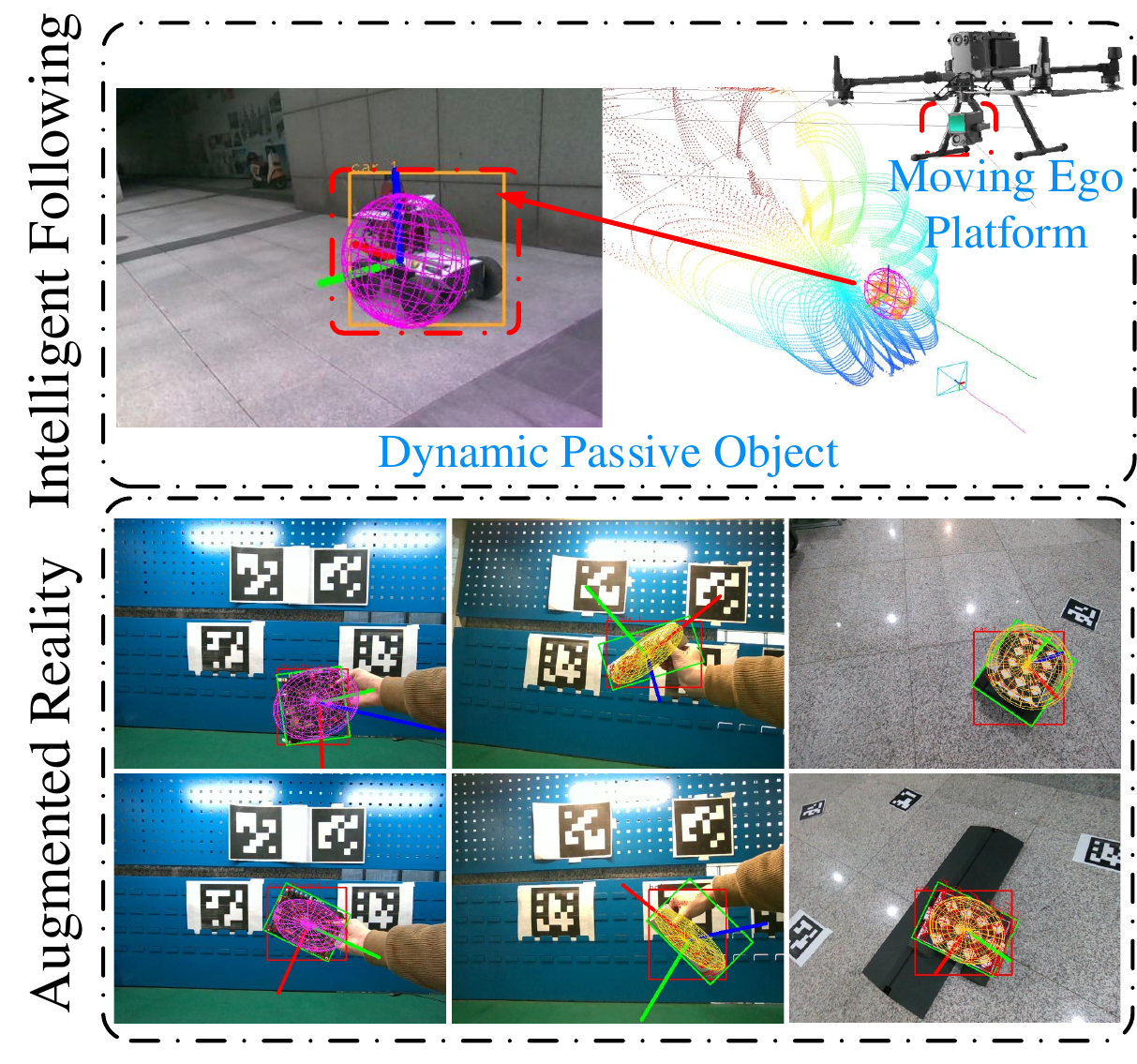}
	\caption{Our proposed method can estimate 6-DoF motion of ego-camera and 9-DoF motion of other unknown rigid objects simultaneously that have potential requirements in intelligent following or dynamic AR.}
	\label{Front_cover}
\end{figure}

Traditional dynamic SLAM focuses on the issues of achieving robustly ego-localization in dynamic scenes and the modeling of static environments. It treats dynamic objects as outliers and remove them directly. 
Compare to these methods, SLAMMOT technology models dynamic objects in the scene into the SLAM system through geometric observations, enabling simultaneous perception of ego-motion and the motion of surrounding objects. \textbf{However, existing methods \cite{16,13,23,26,29,60,28,30,31,eckenhoff2020schmidt,bescos2021dynaslam,qiu2022airdos} focus solely on perceiving object poses and motion states, neglecting the estimation of object scale.} Some approaches \cite{8,9,10,12} directly utilize 3D object detection techniques and deep learning networks to recover object scale, but such methods struggle with scale estimation for unknown objects in general scenes. Other approaches \cite{31,47,liu2021switching} employ simple bounding box fitting methods to estimate object scale, but they are susceptible to noise in the point cloud data and struggle to achieve robust and accurate scale estimation in complex scenes.

Compared to SLAMMOT, Object-Level SLAM focuses on higher-level environmental modeling and representation. It is capable of achieving hierarchical map representation of scenes beyond primitive features, making it suitable for more advanced downstream applications.
Object-Level SLAM technology utilizes semantic and geometric observations to estimate the poses and scales of static objects in the scene. Some works \cite{slam++,nodeslam} employ prior models to represent objects and register and update the modeling results into the global map. Others \cite{liao2022so,26,29,wu2020eao,quadricslam,wang2023qiso} combine semantic observations with multi-view approaches to achieve more generalizable and light-weight representation of unknown object priors through 2D observations. Some of them use cube \cite{26,29,wu2020eao} to characterize objects and optimize object states through sampled projections. Compared to cubes, some others \cite{quadricslam,cao2022object,tian2021accurate,oaslam,liao2022so,wang2023qiso} based on quadric possesses more compact mathematical models and a more complete formulation of projective geometry, enabling nonlinear optimization for object state optimization. \textbf{However, existing works can only model static objects and are unsuitable for dynamic objects. Moreover, they suffer from inaccurate and non-robust initialization under limited viewing angles, leading to erroneous optimization results.}

To address the identified challenges, we propose a unified framework  that concurrently handles ego-motion tracking, 3D tracking of unknown rigid objects, and light-weight modeling using quadric representations for both static and dynamic objects. Table.\ref{Comparison} highlights the distinctions from existing solutions.

\begin{table}[h]\tablefontsmall
\begin{center}
\caption{Comparison with SLAMMOT and Object-Level framework. \textbf{EMT}: Ego Motion Tracking, \textbf{MOT}: Moving Object 3D Tracking, \textbf{SOM}: Static Object Modeling, \textbf{DOM}: Dynamic Object Modeling.}\label{Comparison}
\setlength{\belowcaptionskip}{-1.5cm}   
\resizebox{\linewidth}{!}{
\begin{tabular}{c|c c c c}
\hline
\rowcolor[HTML]{E6E6E6}
\ & \textbf{EMT} & \textbf{MOT} & \textbf{SOM} & \textbf{DOM}\\
\hline
SLAMMOT & $\checkmark$ & $\checkmark$ & $\times$ & $\times$\\
\rowcolor[HTML]{F5F5F5}
Object-Level SLAM & $\checkmark$ & $\times$ & $\checkmark$ & $\times$\\
Ours & $\checkmark$ & $\checkmark$ & $\checkmark$ & $\checkmark$\\
\hline
\end{tabular}
}
\end{center}
\end{table}

\textbf{The main contributions of this paper are the following:}

\begin{itemize}
\item SAM has been integrated into AOT and is capable of accomplishing near real-time detection and tracking of unknown objects, guided by various predefined tasks and prompts.
\item A novel object-centric quadric parameterization is proposed to unify the modeling of static and dynamic objects in the scene. Additionally, we propose a tightly coupled dual-sliding window optimization framework that leverages both semantic and geometric information, enabling us to achieve precise 9 degrees of freedom (9-DoF) estimations for rigid objects.
\item We propose the UniQuadric, which extends the SLAMMOT system for 3D tracking and light-weight modeling of unknown rigid objects, while simultaneously providing ego-localization. Additionally, our system supports both visual and visual-LiDAR fusion configurations, making it suitable for indoor and outdoor scenes.
\end{itemize}
\vspace{-1em}

\section{\bfseries RELATED WORK}
It is crucial to accurately perceive the motion of surrounding objects while achieving ego positioning in augmented reality, autonomous driving and other applications. 
In contrast to traditional dynamic SLAM, Wang et al. \cite{13} presented a system named SLAMMOT, which incorporates dynamic object state estimation into the SLAM framework, enabling the simultaneous estimation of ego-motion and the motion of surrounding rigid moving objects.
Apart from perceiving object motion, accurately estimating the scale of common objects holds significant importance in these applications. Object-Level SLAM has arisen as a pertinent technique for representing the static environment at an object-specific level. In this regard, we will provide a concise overview of the investigations carried out in the realms of SLAMMOT and Object-Level SLAM.
\vspace{-1em}
\subsection{\bfseries SLAMMOT}
Recently, the effectiveness of combining temporal and spatial information in improving the accuracy of object tracking and localization has been verified by Li et al. \cite{26} those who tightly-couples semantic and geometric measurements into a optimization framework to estimate the object's state. 
To improve the quality of object feature data association, VDO-SLAM \cite{28} combines instance segmentation and dense scene flow to achieve feature association, enabling accurate estimation of object pose and velocity. However, the use of multiple networks make it too heavy to satisfy the real-time performance.
Qiu et al.\cite{24} introduced an affordable SLAMMOT solution that fuses monocular camera and IMU data. This approach addresses scale ambiguity across different motion scenarios, leading to precise monocular range determination and object pose estimation. Nevertheless, it doesn't tackle the challenge of object scale estimation.
ClusterSLAM \cite{30} serves as a backend optimization module that implements a scene prior-free algorithm for rigid object detection and motion estimation based on motion consistency constraints. Furthermore, ClusterVO \cite{31} proposes a more comprehensive system that combines object detection and heterogeneous conditional random fields to achieve robust and precise association of object and feature data.
In a recent study, DymSLAM \cite{wang2020dymslam} tackles broader scenarios by employing motion segmentation algorithms to extract masks of unknown objects, enabling pose tracking and point cloud reconstruction for them.
Considering the a prior constraints of the scenario, TwistSLAM\cite{46} uses mechanical joint constraints to restrict the degrees of freedom of an object's pose estimation for specific scenarios, demonstrating the effectiveness of their novel formulation.
\vspace{-1em}
\begin{figure*}[t]
\setlength{\abovecaptionskip}{0.cm}
\setlength{\belowcaptionskip}{-0.5cm}
\setcounter{figure}{1}
\centering
	\includegraphics[width=1.0\linewidth]{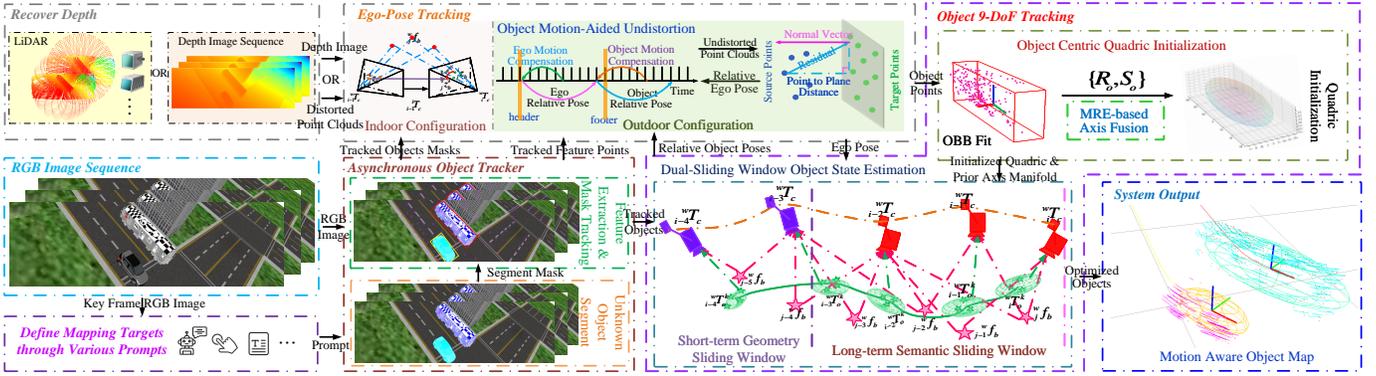}
	\caption{The architecture of proposed system.}
	\label{overview}
\end{figure*}
\subsection{\bfseries Object-Level SLAM} 
The aforementioned researches focus on achieving more precise object pose tracking rather than scale estimation. Compared to it, the objective of Object-Level SLAM is to estimate the poses and scales of objects simultaneously, aiming to create a precise and complete static Object-Level map. SLAM++ \cite{slam++} represents objects in the environment using prior CAD models and adjusts their poses based on multiple frame observations, pioneering the use of object representation in mapping. NodeSLAM \cite{nodeslam} continuously refines the objects in the scene using prior CAD models and applies the idea of Object-Level modeling to the task of grasping. DSP-SLAM \cite{dspslam} has weakened the reliance on prior CAD models by incorporating Signed Distance Function (SDF) and geometric observations, thereby proposing a progressive object SDF model reconstruction system for static scenes, from coarse to fine. To achieve a more lightweight representation of the environment, CubeSLAM \cite{29} represents objects as cubes and recovers object scales using vanishing points combined with multi-view observations. However, the cube-based representation is not compact in parameterization, making it dependent on sampling adjustments during backend optimization. QuadricSLAM \cite{quadricslam} introduces the use of quadric as a representation for objects, using 2D bounding box as semantic constraints. Compared to the cube representation, the compact mathematical model of quadrics allows for nonlinear optimization based on gradient descent in the backend to achieve object modeling. However, its initialization process requires a significant number of frame observations and lacks robustness. OA-SLAM \cite{oaslam} utilized monocular images as input and reparameterized the detection box constraints using a more robust probability distribution. In addition, a quadric initialization method based on semantic observations and triangulation-based depth recovery strategy is developed. Cao et al. \cite{cao2022object} utilize an RGB-D camera and construct convex hull constraints, achieving single-frame robust initialization for most objects. Rui et al. \cite{tian2021accurate} use a stereo camera and incorporate prior axis lengths, enabling fast and robust initialization for specific objects with known axis lengths. And extend the strategy of quadric representation of static landmarks to the outdoor scenes. In our previous work\cite{wu2023object}, we employed both cube and quadric representations to model the scene, applying Object-Level SLAM to applications such as grasping, visual relocalization, and augmented reality in static scenes. However, all of them are only applicable to static scene reconstruction and cannot be applied to dynamic object modeling.

\vspace{-1.5em}
\subsection{\bfseries Unknown Scene Perception} 
In the field of research on object detection and perception for unknown objects, the introduction of Grounding-DINO \cite{liu2023grounding} allows us to perceive objects based on text guidance in an open world scenario. Additionally, the introduction of SAM \cite{SAM} sheds new light on unknown object segmentation, enabling us to input various prompts to determine the objects to be tracked. The recent implementation of MobileSAM \cite{mobilesam} further enhances its practical deployment capabilities, making it more useful in real-world scenarios.

While significant progress has been made in existing methods for SLAMMOT and Object-Level SLAM, certain limitations still endure. No existing system is capable of simultaneously tracking and modeling the poses of both static and dynamic objects in dynamic scenarios. Current quadric modeling methods are effective only for static objects and cannot be applied to dynamic object modeling. Moreover, current solutions are restricted to modeling specific objects and cannot be extended to unknown objects.

For the aforementioned challenges, we introduce a unified framework for dynamic and static object motion estimation and modeling. By integrating SAM, we couple high-level semantic information with low-level geometric information within a dual sliding window framework to simultaneously perform 3D motion tracking and lightweight quadric modeling for unknown rigid objects in the environment.

\section{\bfseries System Overview}

\subsection{\bfseries System Architecture}
Fig.\ref{2} depicts our system framework, which can handle either RGB-D data for indoor scenes or a combination of monocular images and solid-state LiDAR data for outdoor scenes. After defining the task objectives, our system employs the \emph{\textbf{AOT}} module, guided by various prompts, to perform unknown object segmentation and tracking. The resulting tracking masks and associated visual feature points are then input into the \emph{\textbf{Ego-Pose Tracking}} module for ego motion tracking.
This module conducts ego-pose tracking for indoor and outdoor environments based on different configurations and provides object masks, ego-pose, and object points to the \emph{\textbf{Object 9-DoF Tracking}} module. Within the Object 9-DoF Tracking module, we initiate object-centric quadric estimation for the tracked object and refine it by integrating geometric and semantic information using a dual sliding window. Finally, we export the optimized object to create a motion-aware object map.
\vspace{-1.5em}
\begin{figure}[h]
\setlength{\abovecaptionskip}{0.cm}
\setlength{\belowcaptionskip}{-0.2cm}
\setcounter{figure}{2}
	\centering
	\includegraphics[height = 5.5cm,width=1.0\linewidth]{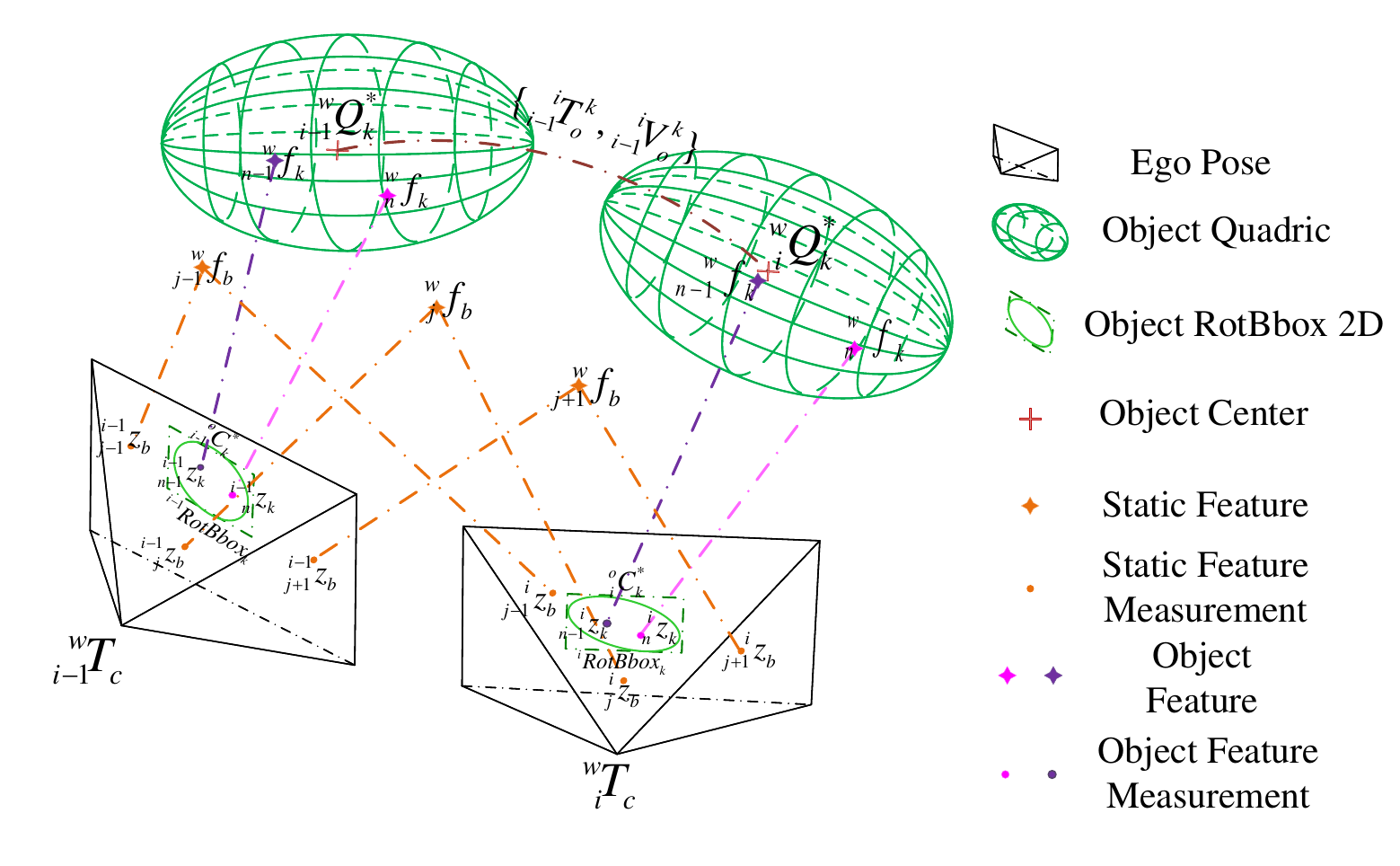}
	\caption{Notation visualization.}
	\label{Notation}
\end{figure}

\subsection{\bfseries Notation}
In this section, we describe the proposed system in detail. The notations used in this paper are as follows and which are visualized as Fig.\ref{Notation}:
\begin{itemize}
	\item ${}_{i}^{w}T_c\in\mathbb{SE}(3)$ - The camera pose of the $i^{th}$ frame in the world frame, which is composed of a camera rotation ${}_{i}^{w}R_c\in\mathbb{SO}(3)$ and a translation $\boldsymbol{}_{i}^{w}t_c\in\mathbb{R}^{3\times1}$.
    \item ${}_{j}^{w}f_b\in\mathbb{R}^{3\times1}$ - The $j^{th}$ background feature position in the world frame. ${}_{j}^{i}z_b\in\mathbb{R}^{2\times1}$ - The $j^{th}$ background feature observation in the $i^{th}$ pixel frame.
    \item ${}_{n}^{o}f_k\in\mathbb{R}^{3\times1}$ - The $n^{th}$ object feature position in the $k^{th}$ object frame. ${}_{n}^{i}z_k\in\mathbb{R}^{2\times1}$ - The $n^{th}$ object feature observation of the $k^{th}$ object in the $i^{th}$ pixel frame. 
	\item $\{_{i-1}^{i}T_{o}^{k},_{i-1}^{i}V_{o}^{k}\}$ - The relative pose and velocity of the $k^{th}$ object between frame $i-1$ and $i$.
    \item $^i{RotBbox}_k=\left\{x_c, y_c, a, b, \theta\right\}$ - The 2D object rotation bounding box (RotBbox) of $k^{th}$ object at $i^{th}$ frame, including axis length, RotBbox center and rotation angle.
    \item $q=[a_x, a_y, a_z, t_x, t_y, t_z, \theta_x, \theta_y, \theta_z]^T \in \mathbb{R}^{9\times1}$ - The quadric parameters of the $k^{th}$ quadric ${^w_i\mathbf{Q}_{k}}$ at frame $i^{th}$ in the world frame, including semi-axis length, translation, rotation. The dual quadric is denoted by ${^w_i\mathbf{Q}_{k}^{*}}\in \mathbb{R}^{4\times4}$.
    \item $^{o}_{i}\mathbf{C}^*_{k}$ - The projection dual conic of the dual quadric ${^w_{i}\mathbf{Q}_{k}^{*}}$ of $k^{th}$ object at the $i^{th}$ frame. 
\end{itemize}

\section{\bfseries Asynchronous Object Tracker}\label{ASYNCTracker_SEC}
Accurate detection and stable tracking of the target object are fundamental for implementing backend batch optimization. Grounding-DINO, SAM, and DeAOT\cite{DeAOT} have demonstrated impressive capabilities in unknown object tracking and segmentation. However, the detection process remains time-consuming. To meet practical application demands, we introduce the AOT (shown in Fig.\ref{ASYNCTracker}) designed for unknown object detection and tracking. This tracker operates with two threads:
the detection thread detects and segments target objects in keyframes using various prompts. Where the $\Delta t_{segment}$ is the time for each segmentation. After that, the tracking thread will conducts pixel-level tracking for $SegMask_{i}$. After receiving the segmentation results, the tracking thread first performs a \emph{Backward Association} operation to initialize the new object or update the tracked objects' mask, ensuring synchronization of timestamps between the detection and tracking threads. Then, it employs \emph{Jump Track} operation to propagate the updated masks to the current timestamp to ensure continuity in subsequent tracking. Here, $\Delta t_{track}$ represents the time for each tracking step.

Moreover, for each incoming frame, the object features association strategies are followed:

1) For each 2D RotBbox, we extract fast corners within the corresponding object's bounding box area, identified by its unique ID. We then establish inter-frame correlation by applying the Kanade-Lucas-Tomasi (KLT) optical flow algorithm. Additionally, for objects with pre-estimated poses, we leverage prior knowledge of their motion to enhance feature association. This approach helps alleviate the problem of erroneous feature associations caused by the coupling between object and ego-motion.   

2) Benefiting from the envelope of the quadric, we can roughly filter out foreground and background features by applying a distance threshold within the initialized ellipsoid. Feature points located outside the axis length of the quadric are classified as background features, while those falling within the envelope are selected for subsequent object tracking and motion estimation.

\vspace{-0.5em}

\section{\bfseries STATE ESTIMATION}\label{state_estimation}
The proposed system aims to estimate the state of ego and surrounding objects states simultaneously. The object state estimation factor graph is shown as Fig.\ref{FactorGraph} which integrates hybrid observations in a dual sliding window to achieve joint multi-states estimation.
\vspace{-1.5em}
\begin{figure}[t]
\setlength{\abovecaptionskip}{0.cm}
\setlength{\belowcaptionskip}{-0.5cm}
\setcounter{figure}{3}
	\centering
	\includegraphics[width=1.0\linewidth]{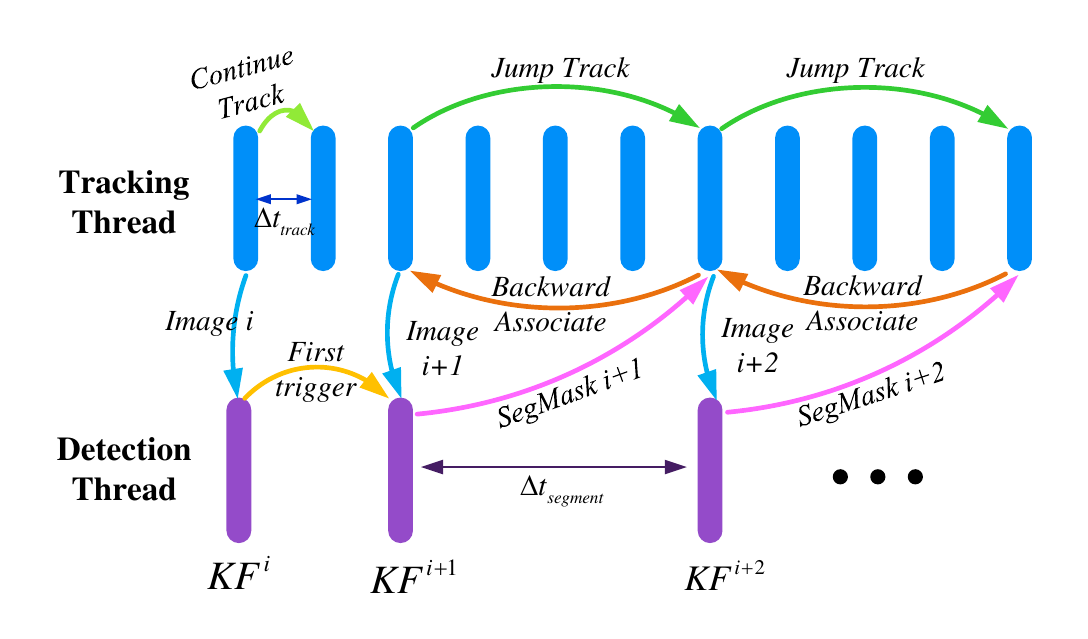}
	\caption{Asynchronous Object Tracker.}
	\label{ASYNCTracker}
\end{figure}
\subsection{\bfseries Ego-motion Estimation}
\subsubsection{\bfseries Indoor ego state estimation}\label{Indoor_ego_state_estimation}
In indoor scene applications, we utilize RGB-D cameras as sensor input and perform camera pose tracking based on ORB-SLAM3\cite{campos2021orb}. In contrast to the original system, we incorporate semantic information to assist in decoupling state estimation, enabling a more comprehensive description of various landmarks in the scene. The Maximum A Posteriori Estimation (MAP) problem, which constructs visual observations used for pose estimation, can be represented as:
\begin{equation}
\begin{split}
\begin{aligned}
&{}^{w}T_c,{}^{w}T_o,{f_{label}} = \\
&{\rm{           }}\mathop {\arg \max }\limits_{{}^{w}T_c,{}^{w}T_o,{f_{label}}} \mathop \Pi \limits_{i = 0}^N \mathop \Pi \limits_{j = 0}^{{M_{label}}}p(_{j}^{i}z_{label}|{}_{i}^{w}T_c,{}^{w}T_o^{label},_{j}^{w}f_{label})\label{1}
\end{aligned}
\end{split},
\end{equation}
where, $(\cdot)_{label}$  represents the semantic label of the landmark, $N$ and $M_{label}$ represent the number of frames and landmarks, respectively. When estimating its own pose, we utilize landmarks with a semantic label of \emph{background} as observations and $^{w}_{i-1}T_o^{b} = ^{w}_{i}T_o^{b}$, specifically, we make the assumption that all stationary landmarks are situated on a fixed, immobile rigid body. As a result, the pose of this rigid body remains constant across all frames.

\subsubsection{\bfseries Outdoor ego state estimation}\label{outdoor_ego_state_estimation}
To overcome the constraints of RGB-D cameras, which may not provide precise depth measurements in outdoor environments, we adopt a solid-state LiDAR sensor in conjunction with monocular camera as input sensors for the subsequent state estimation in outdoor scenarios.

In the process of ego-pose estimation, when dealing with each incoming LiDAR scan, we account for motion distortions induced by ego-motion within the frame and dynamic objects by applying a motion model for compensation. Subsequently, we employ an error state iterated Kalman filter (ESIKF) that focuses on minimizing point-to-plane residuals to estimate the system's state, as detailed in \cite{xu2022fast}.
In our sensor fusion configuration for object state estimation, we leverage LiDAR point clouds to provide depth information for visual features, while RGB images serve for inter-frame feature association and semantic extraction. To recover feature depth, we employ spherical projection in conjunction with K-Nearest Neighbor search (KNNs) to establish data association between LiDAR point clouds and visual features. We further refine feature depth through line-of-sight interpolation. 

To address the synchronization issue between LiDAR and camera frame rates, we employ a soft synchronization strategy for timestamp alignment.

Additionally, when estimating the state of moving objects, the distortion present in LiDAR point clouds on these objects is a result of the coupling between their own motion and the motion of the objects. Utilizing existing distortion correction strategies designed for static scenes are ineffective in eliminating this distortion.
To address this challenge, we integrate the estimated object motion and introduce a motion-aware compensation strategy. The algorithm diagram is illustrated in Fig. \ref{motion_compensation}.
First, the $n^{th}$ raw LiDAR point ${ }^t_{n} \tilde{p}$ with timestamp $t$ in LiDAR frame can be undistorted to the current world frame by using ego-motion through:  
\begin{equation}
\begin{aligned}
{ }_n^t p={R_{t}} {^t_n\tilde{p}}+t_t=R_{t_s}\left(R_{t_s t}{{ }^t_{n} \tilde{p}}+t_{t_s t}\right)+t_{t_s}
\label{2}
\end{aligned},
\end{equation}

where $R_{t_s t}, t_{t_s t}$ can be obtained through quaternion spherical interpolation and linear interpolation methods as:
\begin{equation}
\begin{aligned}
\left\{\begin{array}{l}
R_{t_s t}=\frac{\sin \left(\left(1-\frac{t-t_s}{t_e-t_s}\right) \cdot{ }_{t_s}^{t_e} \theta\right){ }_{t_s}^w q+\sin \left(\frac{t-t_s}{t_e-t_s} \cdot_{t_s}^{t_e} \theta\right){ }_{t_e}^{w} q}{\sin \left({ }_{t_s}^{t_e} \theta\right)} \\
t_{t_{s t}}=\frac{t-t_s}{t_e-t_s} \cdot t_{t_s t_e}
\end{array}\right.
\label{3}
\end{aligned}
\end{equation}
\begin{equation}
\begin{aligned}
{ }_n^t p^k=R_{t_s t}^k {^t_{n}p}+t_{t_s t}^k
\label{4}
\end{aligned}
\end{equation}
\vspace{-2.0em}
\begin{figure}[h]
\setlength{\abovecaptionskip}{0.cm}
\setlength{\belowcaptionskip}{-0.5cm}
\setcounter{figure}{4}
	\centering
\includegraphics[width=1.0\linewidth]{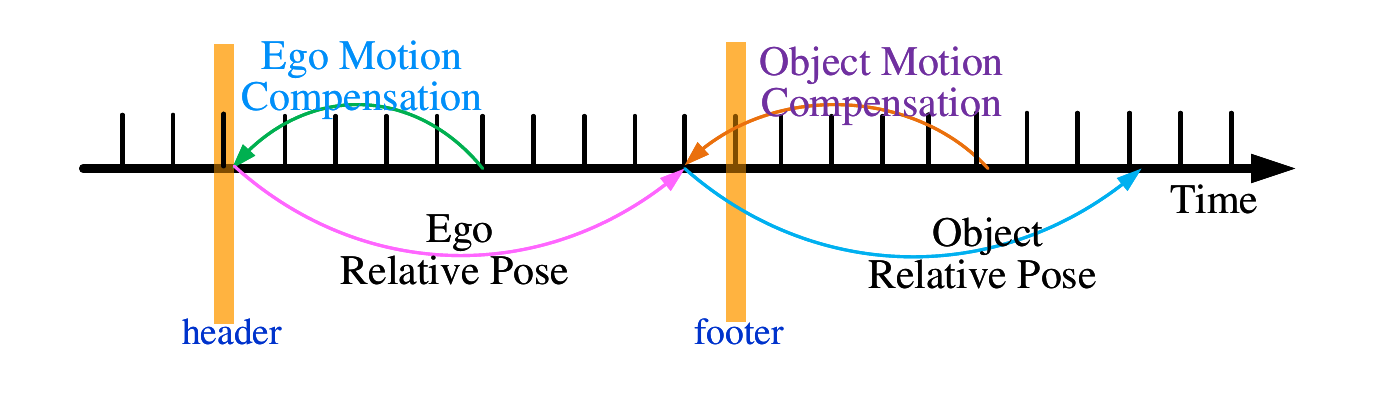}
	\caption{Motion Compensation.}
	\label{motion_compensation}
\end{figure}

Furthermore, we utilize the estimated object motion to obtain undistorted point clouds using Eq.\ref{4}, which will be utilized in the subsequent estimation of the object's state.
\begin{figure}[t]
\setlength{\abovecaptionskip}{-0.2cm}
\setlength{\belowcaptionskip}{-0.5cm}
\setcounter{figure}{5}
	\centering
	\includegraphics[width=1.0\linewidth]{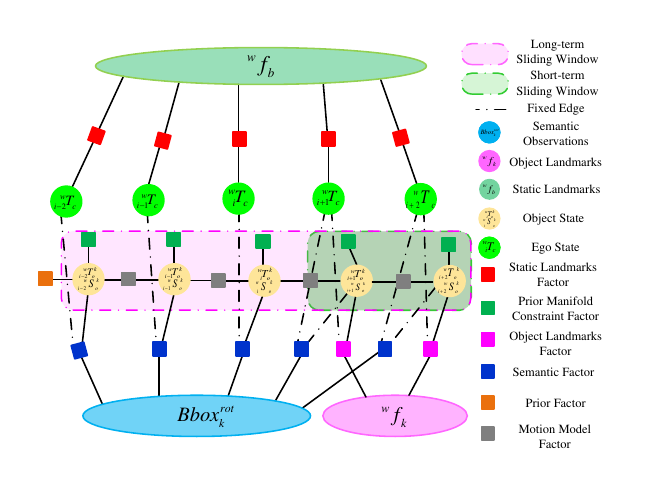}
	\caption{Object State Estimation Factor Graph.}
	\label{FactorGraph}
\end{figure}
\vspace{-1.5em}
\subsection{\bfseries Object-state Estimation}\label{object_state}
\vspace{-0.5em}
To estimate the pose and scale of rigid objects within the environment, we introduce a dual sliding window optimization framework which combines long-term and short-term parts to fuse multiple sources of information. The novel object state estimation factor graph shown as Fig.\ref{FactorGraph} which can be formulized as:
\begin{equation}
\begin{aligned}
& { }^w T_o^k,S_o^k,{ }^o f^k=\underset{{ }^{w} T_c,{ }^{w} T_o^k,S_o^k,{ }^o f^k}{\arg \min }\{{e_{\text {prior }}}+\\
& \sum_{i=0}^{N_k} \sum_{n=0}^{M_o^k}\left\|e_z\left({ }_n^i z^k \mid{ }_i^w T_c,{ }_{i-1}^w T_c,{ }_i^w T_o,{ }_{i-1}^w T_o^k,{ }_n^o f^k\right)\right\|_{\Sigma_n}^2+ \\
& \sum_{i=1}^{N_k}\left\|e_M\left({ }_i^w T_o^k{ }_o^w{ }_{i-1} T_o^k\right)\right\|_{\Sigma_k^i}^2+\sum_{i=0}^{N_k}\left\|e_{\mathcal{B}}\left(_i{B}^k, \mathcal{Q}^k\right)\right\|_{\Sigma_o^i}^2\\
& +\sum_{i=0}^{N_k}\left\|e_{\mathcal{A}}\left(_i{A}^k, _i{\tilde{A}}^k\right)\right\|_{\Sigma_A^i}^2+\sum_{i=0}^{N_k}\left\|e_{\mathcal{C}}\left(_i{C}^k, \tilde{C}_i\right)\right\|_{_i{\Sigma_C}^k}^2\}
\label{5}
\end{aligned}
\end{equation}

In our implementation, we manage a short-term sliding window containing 8-10 keyframes. This window serves to incorporate the geometric constraint and motion model constraint for object pose estimation. 
In the long-term sliding window which contains 20-25 keyframes, we incorporate semantic observations of objects to constrain their states. Differing from conventional geometric features, semantic information can offer more enduring observations, thereby providing more consistent constraints. Additionally, to prevent convergence issues resulting from inaccurate initial pose estimation when integrating semantic constraints, we exclusively include the keyframes that have been refined within the short-term sliding window into the long-term sliding window. This selection is based on the observation that keyframes optimized within the short window exhibit a more reliable initial state, which in turn guarantees the convergence of the long-term window. Additionally, as we mainly focus on object state estimation, when estimating the object's state, we keep the ego-pose fixed and use it solely for coordinate transformations.

\subsubsection{\bfseries Short-term object state estimation}\label{Short-term object state estimation}
\

\textbf{Geometric reprojection factor:} According to the AOT designed in Sec.\ref{ASYNCTracker}, we obtain masks for each tracked object, which help determine the feature points belonging to that object within the masked regions. In the process of modeling the visual geometric constraints, we use the standard geometric reprojection factor\cite{28}\cite{31}.
Firstly, we transform the landmark $^c_{n}f_k^{i-1}$ to the coordinate of the $k^{th}$ object using $^w_{i-1}T_c$ and $^w_{i-1}T_o^k$. Based on the rigid assumption that geometric landmarks belonging to the same object maintain consistent positions relative to the object's coordinates across different frames. With this assumption, we can further transform the landmark $^c_{n}f_k^{i-1}$ to the world coordinate of the $i^{th}$ frame using $^w_{i}T_c$ and $^w_{i}T_o^k$. Subsequently, by applying the reprojection operation $\Omega(\cdot)$, we can project the landmark to the current pixel coordinate. Finally, we establish the object pose constraints by comparing the projected landmark with the associated feature point $^i_{n}z_k$ as follows:
\begin{equation}
\begin{aligned}
& e_z\left({ }_n^i z^k \mid{ }_i^w T_c,{ }_{i-1}^w T_c,{ }_i^w T_o,{ }_{i-1}^w T_o^k,{ }_n^o f^k\right)=\\
& \Omega\left(\left({ }_i^w T_c\right)^{-1}{ }_i^w T_o^k\left({ }_{i-1}^w T_o^k\right)^{-1}{ }_{i-1}^w T_c{ }^c_n f^{i-1}_k\right) - { }_n^i z^k
\label{6}
\end{aligned}
\end{equation}

\textbf{Motion model factor:} To address the problem of trajectory jumps caused by visual observation noise, we incorporate motion model factor $e_M$ into the optimization process to smooth the 3D trajectory. The motion model factor can be formulized as:
\begin{equation}
\begin{aligned}
{e_M}({}_{i}^{w}T_o^k,{}_{i-1}^{w}T_{o}^{k}) = {}_{i}^{w}T_{o}^{k} - {}_{i}^{w}\hat T_{o}^{k},\label{7}
\end{aligned}
\end{equation}
where ${}^wT_o^{k,i}$ is the observation of the $k^{th}$ object in frame $i$, and ${}^w\hat T_o^{k,i}$ is the predicted pose of the object in frame $i-1$ obtained using the motion model.

\subsubsection{\bfseries Long-term object state estimation}\label{Long-term object state estimation}
\

\textbf{Object Centric Quadric Initialization:} In this section, we present a mathematical analysis of the exist dual quadric formulation to illustrate the limitations of it for representing dynamic objects. The quadratic represented in the world frame can be given by:
\begin{equation}
\begin{aligned}  
{^{w}_{i}{\mathbf{Q}}^*} = \begin{bmatrix}
    ^w_i{R} & ^w_i{t} \\
    0^{T} & 1
    \end{bmatrix}
    \begin{bmatrix}
    D & 0 \\
    0^{T} & -1
    \end{bmatrix}
    \begin{bmatrix}
    ^w_i{R}^T & 0 \\
    ^w_i{t}^{T} & 1
    \end{bmatrix}\\
\end{aligned},
\label{8}
\end{equation}
where the dual quadric is denoted by ${^{w}_{i}{\mathbf{Q}}^*}\in \mathbb{R}^{4\times4}$, $D\in{R^{3\times3}}$ is the diagonal matrix composed of the squares of the quadric axis lengths, $^w_i{R}\in{R^{3\times3}}$ and $^w_i{t}\in{R^{3\times1}}$ is the quadric centroid translation and rotation in the world frame respectively. And the projection dual conic denoted by $^{w}_i\mathbf{C}^*$, 
\begin{equation}
\begin{aligned} 
    ^{w}_i\mathbf{C}^*=K ^{c}_{i}{T}_{w}{^{w}_{i}{\mathbf{Q}}^*}({K ^{c}_{i}{T}_{w}})^T
\end{aligned}
\label{9}
\end{equation}
\vspace{-1.5em}

The above formulation is based on the static environment assumption, it's means that the pose of an object in the world frame is fixed and independent of time. This limitation prevents its applicability to dynamic object representation. To address this problem, we based on rigid assumption and propose an \emph{"object-centric quadric formulation"} for both dynamic and static objects. We assume that the pose of the quadric ${\mathbf{Q}}^*$ in the object frame is constant. And the expression for the object-centric quadric ${^o\mathbf{Q}^*}$ can be reformulized as:
\vspace{-0.5em}
\begin{equation}
\begin{aligned}
^{o}\mathbf{Q}^* = \begin{bmatrix}
    \mathbf{I} & \mathbf{0} \\
    \mathbf{0}^{T} & 1
    \end{bmatrix}
    \begin{bmatrix}
    \mathbf{D} & \mathbf{0} \\
    \mathbf{0}^{T} & -1
    \end{bmatrix}
    \begin{bmatrix}
    \mathbf{I}^T & \mathbf{0}^{T} \\
    \mathbf{0} & 1
    \end{bmatrix}\\
    \end{aligned},
\label{10}
\end{equation}
where we initialize the rotation matrix $R$ of the object in the object coordinate as the identity matrix, and the center of the quadirc is located at the origin of the object coordinate. Furthermore, the new projection dual conic $^{o}_{i}\mathbf{C}^*$ of the quadric ${^o\mathbf{Q}^*}$ at the $i^{th}$ frame can be described as follows:
\begin{equation}
\begin{aligned} 
    ^{o}_{i}\mathbf{C}^*={K {^{c}_{i}T_{w}}}  {^{w}_{i}T_{o}}{^{o}_{i}\mathbf{Q}^*}({K {^{c}_{i}T_{w}} {^{w}_{i}T_{o}}})^T
\end{aligned}
\label{11}
\end{equation}
\vspace{-1em}
\begin{algorithm}[H]
  \renewcommand{\algorithmicrequire}{\textbf{Input:}}
  \renewcommand{\algorithmicensure}{\textbf{Output:}}
  \caption{Oriented Bounding Box Fit based on RANSAC}\label{OBBRANSAC}
  \begin{algorithmic}[1]
  \setstretch{1.2}
  \Require $^i\mathbb{P}^k$ - The point cloud of $k^{th}$ object at $i^{th}$ frame, $Num$ - The maximum iterations, $\epsilon$ - Inliers threshold.
  \Ensure $^i{obb}_{best}^k$ - Best oriented bounding box, $^i{\delta}_{axis}^k$ - Axis length uncertainty.
  \State \textbf{Initialize:} $Inlier_{best}$ - Inlier set, $\mathbb{A}$ - Axis length set.
  \State \textbf{for} $t=1;t \le Num;n++$ \textbf{do}
  \State \quad \quad $Sample \leftarrow$ RandomSample($^i\mathbb{P}^k$)
  \State \quad \quad $obb \leftarrow$ FitOrientedBoundingBox($Sample$)
  \State \quad \quad $Inliers \leftarrow$ ComputeInliers($^i\mathbb{P}^k,obb,\epsilon$) \Comment{Eq.(\ref{12})}
  \State \quad \quad \textbf{if} $len(Inliers) > len(Inlier_{best})$ \textbf{then}
  \State \quad \quad \quad $^i{obb}_{best}^k = obb$
  \State \quad \quad \quad $Inlier_{best} = Inliers$
  \State \quad \quad \quad $\mathbb{A}.append(obb.axis)$
  \State \quad \quad \textbf{end if}
  \State \textbf{end for}
  \State $^i{\mu}_{axis}^k=mean(\mathbb{A})$,$^i{\delta}_{axis}^k=\frac{\sum_{n=0}^{len(\mathbb{A})}(\mathbb{A}^n-^i{\mu}_{axis}^k)}{len(\mathbb{A})}$
  \State $^i{obb}_{best}^k.axis=(1-\omega)\cdot^i{\mu}_{axis}^k+\omega\cdot^i{\delta}_{axis}^k$
  \end{algorithmic}
\end{algorithm}
To robustly initialize the quadric, we propose the scale-constrained quadric initialization strategy (SQI) in which the quadric is first initialized in a sphere and then refined in the form of a scale-constrained quadric as more 2D detections are observed. The quadric’s axis length and object orientation can be initialized by OBB that construct by object point clouds. The procedure process of fitting OBB in our method is illustrated on Alg.\ref{OBBRANSAC}.In order to enhance the fitting robustness, we implemented the algorithm within the RANSAC framework. 
\begin{equation}
\begin{aligned} 
\|(p-obb.center)\cdot obb.axis-obb.axis\|_2
\end{aligned}
\label{12}
\end{equation}
where, the Eq.(\ref{12}) is used to evaluate the inliers of the model fitting at each iteration. Additionally, during the fitting process, we assessed the uncertainty of the fitting results based on the stability observed throughout the iterations, which provides a fusion prior for subsequent multi-frame fusion.
\begin{figure}[h]
\setlength{\abovecaptionskip}{0.cm}
\setlength{\belowcaptionskip}{-0.5cm}
\setcounter{figure}{6}
	\centering
	\includegraphics[width=1.0\linewidth]{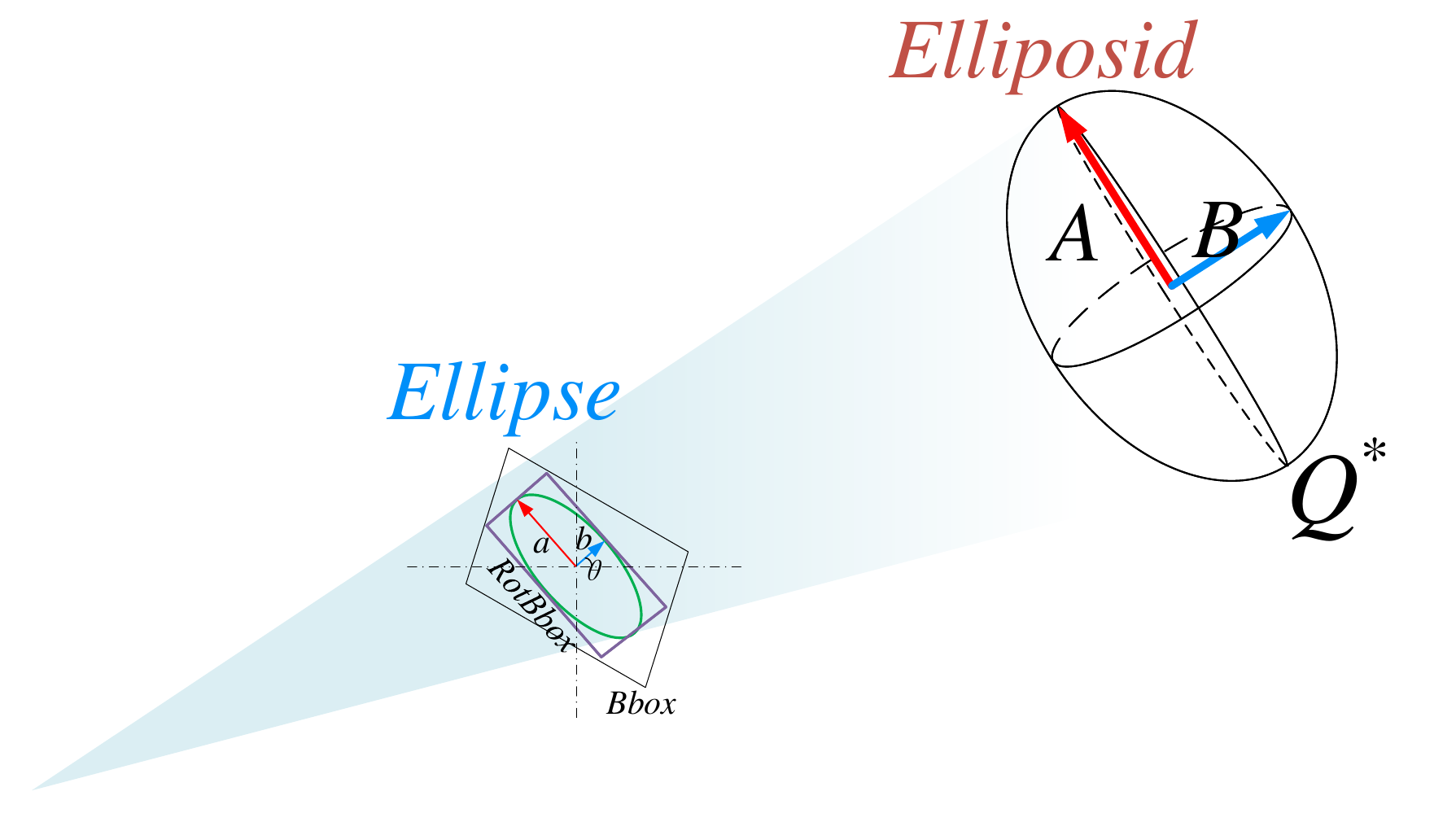}
	\caption{RotBbox Quadric Observation.}
	\label{rot_rect_constrain}
\end{figure}

\textbf{RotBbox observation factor:} The parameters of a quadric can be estimated through observations from multiple views. In our method, we can obtain the objects' RotBbox by leveraging the pixel-level semantic segmentation results provided by SAM. The semantic observation provided by RotBbox is illustrated as Fig.\ref{rot_rect_constrain}, where the RotBbox can be parameterized as $RotBbox_{2D}=\left\{x_c, y_c, a, b, \theta\right\}$. Based on the RotBbox, we can obtain the dual conic observation for the current object $k^{th}$ at time $i$:
\begin{equation}\label{13}
\hspace{-1.05em}
\setlength{\arraycolsep}{1.0pt}
C_{obs}^* = \begin{bmatrix}
\cos \theta & -\sin \theta & x_c \\
\sin \theta & \cos \theta & y_c \\
0 & 0 & 1
\end{bmatrix} \begin{bmatrix}
a^2 & 0 & 0 \\
0 & b^2 & 0 \\
0 & 0 & -1
\end{bmatrix} \begin{bmatrix}
\cos \theta & -\sin \theta & x_c \\
\sin \theta & \cos \theta & y_c \\
0 & 0 & 1
\end{bmatrix}^T
\end{equation}

Similarly, for an ellipsoid, we can obtain its projection dual conic in the current frame based on the Eq.(\ref{11}).

Regarding the representation of rotated rectangles in observations, we represent the dual conic ellipse as a 2D Gaussian distribution, which can be denoted as $\mathcal{N}(\mu,\sigma)$. Here, $\mu = \left(x_c,y_c\right)^T$ represents the center of the ellipse, and sigma represents the covariance obtained through dual conic decomposition. Motivated by \cite{zins2022level}, we utilize the Bhattacharyya distance to quantify the similarity between probability distributions. Thus, we can calculate the semantic observation $e_{\mathcal{B}}$ based on RotBbox by:
\begin{equation}\label{14}
\begin{aligned}
e_{\mathcal{B}}\left({B}_{obs}, \mathcal{Q}_{est}\right)= & e_{\mathcal{B}}\left(\mathcal{N}_{obs}, \mathcal{N}_{est}\right)\\
= & \frac{1}{8}\left(\mu_{e s t}-\mu_{o b s}\right)^T\Sigma^{-1}\left(\mu_{e s t}-\mu_{o b s}\right) \\
& +\frac{1}{2} \ln \left(\frac{\operatorname{det}\left(\Sigma\right)}{\sqrt{\operatorname{det} \Sigma_{e s t} \cdot \operatorname{det} \Sigma_{o b s}}}\right)
\end{aligned},
\end{equation}
where $\Sigma = \frac{\Sigma_{e s t}+\Sigma_{o b s}}{2}$.
\begin{table*}[h]\tablefont
\begin{center}
\caption{Performance comparison on SW4 and O2 sequence in Oxford Multimotion dataset. Best results are highlighted as \sethlcolor{colorh}\hl{first}, \sethlcolor{colorm}\hl{second}, and \sethlcolor{colorl}\hl{third}.}\label{OMD}
\setlength{\belowcaptionskip}{-1.5cm}   
\resizebox{\linewidth}{!}{
\begin{tabular}{c|c c c c|c c c c|c c c c|c c c c}
\hline
\multicolumn{1}{c|}{\cellcolor[HTML]{E6E6E6}} & \multicolumn{4}{c|}{\cellcolor[HTML]{E6E6E6}ClusterVO\cite{31}} & \multicolumn{4}{c|}{\cellcolor[HTML]{E6E6E6}DymSLAM\cite{wang2020dymslam}} & \multicolumn{4}{c|}{\cellcolor[HTML]{E6E6E6}MVO \cite{judd2018multimotion}} & \multicolumn{4}{c}{\cellcolor[HTML]{E6E6E6}Proposed Approach} \\
\rowcolor[HTML]{E6E6E6}
\multirow{-2}{*}{Sequence} & Trans(m) & Roll(°) & Yaw(°) & Pitch(°) & Trans(m) & Roll(°) & Yaw(°) & Pitch(°) & Trans(m) & Roll(°) & Yaw(°) & Pitch(°) & Trans(m) & Roll(°) & Yaw(°) & Pitch(°)\\
\hline
SW4-Ego & \colorl{0.62} & \colorl{-4.97} & \colorl{2.53} & \colorm{0.448} & \colorh{0.04} & \colorm{-0.78} & \colorh{-1.6} & -0.7 & 0.93 & -6.82 & 3.13 & \colorh{0.16} & \colorm{0.29} & \colorh{0.31} & \colorm{1.9} & \colorl{0.51}\\
\rowcolor[HTML]{F5F5F5}
SW4-C1 & \colorl{0.24} & \colorh{0.65} & \colorm{6.05} & \colorh{0.145} & \colorh{0.11} & -5.87 & -9.09 & \colorl{-2.77} & 0.36 & \colorl{3.09} & \colorl{6.51} & \colorm{0.16} & \colorm{0.23} & \colorm{0.75} & \colorh{4.12} & \colorm{0.16}\\
SW4-C2 & \colorl{0.448} & \colorl{23.19} & -62.53 & \colorh{0.99} & \colorm{0.13} & \colorm{7.13} & \colorh{-2.75} & -6.88 & 0.64 & 25.26 & \colorl{-55.83} & \colorl{1.46} & \colorh{0.10} & \colorh{6.05} & \colorm{-3.22} & \colorm{1.08}\\
\rowcolor[HTML]{F5F5F5}
SW4-C3 & \colorl{0.243} & -13.96 & \colorm{0.48} & 5.54 & \colorm{0.16} & \colorm{3.66} & \colorm{-4.65} & \colorh{3.56} & 0.45 & \colorl{-11.35} & \colorl{0.53} & \colorm{4.08} & \colorh{0.11} & \colorh{2.5} & \colorh{0.35} & \colorl{5.13}\\
SW4-C4 & \colorl{4.69} & 243.25 & 23.65 & -101.05 & \colorh{0.05} & \colorh{-3.29} & \colorh{-2.43} & \colorm{1.8} & 5.94 & \colorl{93.56} & \colorl{5.77} & \colorl{-53.75} & \colorm{0.51} & \colorm{6.53} & \colorm{2.8} & \colorh{1.12}\\
\rowcolor[HTML]{F5F5F5}
O2-Ego & \colorm{0.24} & \colorm{1.00} & \colorh{-0.225} & \colorh{0.484} & - & - & - & - & \colorl{0.31} & \colorl{3.45} & \colorl{-3.21} & \colorl{1.73} & \colorh{0.19} & \colorh{0.65} & \colorm{1.1} & \colorm{0.51}\\
O2-Cuboid & \colorm{0.19} & \colorm{1.42} & \colorm{-8.45} & \colorh{0.44} & - & - & - & - & \colorl{0.51} & \colorl{1.94} & \colorl{-22.83} & \colorl{1.75} & \colorh{0.15} & \colorh{0.95} & \colorh{6.77} & \colorm{0.80}\\
\rowcolor[HTML]{F5F5F5}
O2-Cube & \colorh{0.48} & \colorm{19.35} & \colorm{-7.23} & \colorm{12.03} & - & - & - & - & \colorl{1.41} & \colorl{37.22} & \colorl{-27.83} & \colorl{19.72} & \colorm{0.80} & \colorh{10.35} & \colorh{5.13} & \colorh{9.56}\\
\hline
\end{tabular}
}
\end{center}
\end{table*}

\begin{figure*}[t]
\setlength{\abovecaptionskip}{0.cm}
\setlength{\belowcaptionskip}{-0.2cm}
\setcounter{figure}{7}
	\centering
	\includegraphics[width=1.0\linewidth]{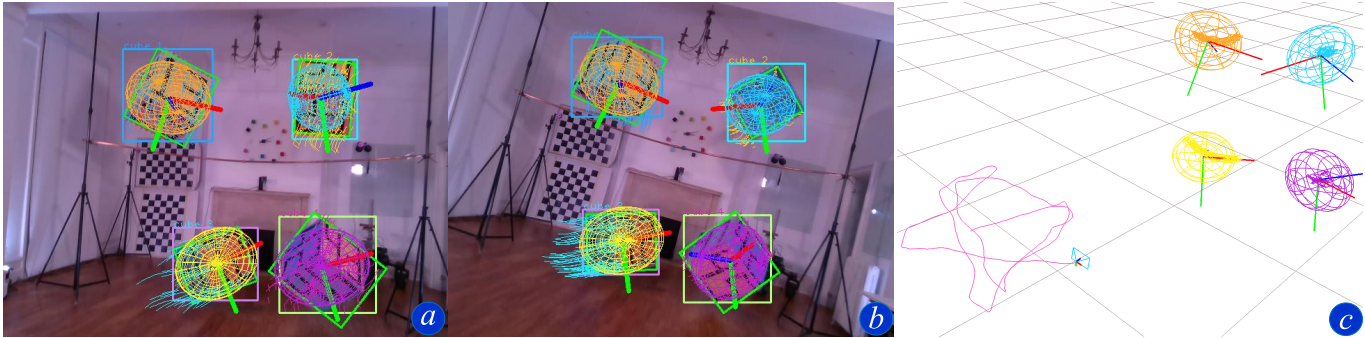}
	\caption{Qualitative results in OMD Sequence SW4. The subfigures demonstrate the proposed method can estimate multi-states simultaneously.}
	\label{Oxford_SW4}
\end{figure*}
\begin{figure*}[t]
\setlength{\abovecaptionskip}{0.cm}
\setlength{\belowcaptionskip}{-0.3cm}
\setcounter{figure}{8}
	\centering
	\includegraphics[width=1.0\linewidth]{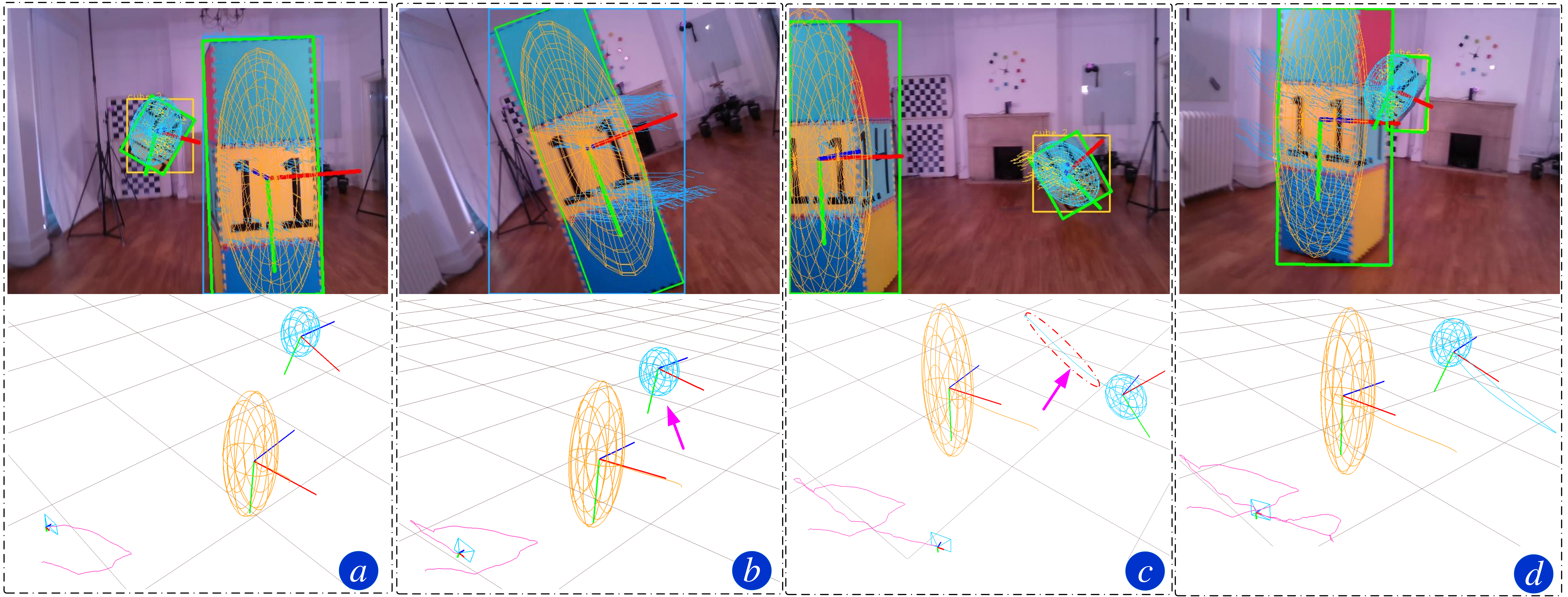}
	\caption{Qualitative results in OMD Sequence O2. The subfigures demonstrate the system's ability to maintain continuous and stable object pose and scale estimations even when the tracked object experiences temporary occlusions. Figures a to d depict scenarios with double occlusions.}
	\label{Oxford_O2}
\end{figure*}
\textbf{Prior axis manifold factor:} 
The axis lengths are important parameters used to describe the shape of a quadric. However, when there are limited observation angles for objects in the scene, relying solely on the 2D constraints from bounding boxes is insufficient for accurate quadric estimation. To overcome the problem, we incorporates prior axis length manifold constraints into the optimization process of quadric to enhance the stability and accuracy of the estimation results.

The more accurate the prior estimation of the axis lengths, the more accurate the constraints will be. In our method, we employ a multi-frame OBB fusion algorithm based on MRE to estimate the prior axis lengths. The algorithm flow is illustrated in Alg.\ref{MREPAE}.
The algorithm is built upon Alg.\ref{OBBRANSAC}. In each iteration of the OBB fitting, we calculate the uncertainty of the fitting result. During the axis length fusion process, we aim to obtain a more accurate estimation of the axis lengths by fusing multiple frames of data. Each frame's estimated axis length results are treated as information, and their weights are determined based on the uncertainty of each frame. The fusion process minimizes the entropy of the resulting fusion by minimizing the overall uncertainty, leading to the most accurate estimation possible.
The residual form based on the prior axis length can be formulized as:
\begin{equation}\label{15}
\begin{aligned}
e_{\mathcal{A}}\left(_i{A}^k, _i{\tilde{A}}^k\right) = _i{A}^k - _i{\tilde{A}}^k
\end{aligned},
\end{equation}
where $_i{A}^k, _i{\tilde{A}}^k$ denote the axis length and prior axis of $k^{th}$ object at $i^{th}$ frame respectively.
\begin{algorithm}[H]
  \renewcommand{\algorithmicrequire}{\textbf{Input:}}
  \renewcommand{\algorithmicensure}{\textbf{Output:}}
  \caption{Minimum Renyi Entropy-based Prior Axis Estimation}\label{MREPAE}
  \begin{algorithmic}[1]
  \setstretch{1.2}
  \Require $\mathbb{A}^k$ - Axis length estimates of $k^{th}$ object for each frame, $\mathbb{\delta}^k$ - Uncertainty values of $k^{th}$ object for each frame, Num - The maximum iterations.
  \Ensure ${A}_{prior}^k$ - The fused prior axis.
  \State \textbf{Initialize:} 
  \State $\mathbb{A}^k_{norm}$ = NormalizeAxis($\mathbb{A}^k$).\Comment{\textbf{Normalize $^i{{A}}^k$ to ensure they are in the same range.}}
  \State $\mathbb{P}^k$ = CalculateProbabilities($\mathbb{\delta}^k$).\Comment{\textbf{For each $i$, compute the probability $^i{{P}^k}=exp(-^i{\delta}^k/2)$.}}
  \State $M = len(\mathbb{A}^k)$.
  \State \textbf{for} $t=1;t \le Num;n++$ \textbf{do}
  \State \quad \quad $^tA_{fused}^k=\frac{1}{1-\alpha} \cdot log(\sum_{i=0}^{M}((^i{p}^k)^{\alpha}\cdot exp(^i{A}^k)))$
  \State \quad \quad $H_{\alpha} = -^tA_{fused}^k$\Comment{\textbf{Calculate the Renyi Entropy}}
  \State \quad \quad \textbf{Optimize $\alpha$ using gradient descent algorithm.}
  \State \textbf{end for}
  \State ${A}_{prior}^k = \frac{1}{1-\alpha} \cdot log(\sum_{i=0}^{M}((^i{p}^k)^{\alpha}\cdot exp(^i{A}^k)))$
  \end{algorithmic}
\end{algorithm}

\textbf{Rigid prior factor:}\label{Rigid_prior_factor} The lack of multiple-view observations for the tracked object in the scene can result in under-constrained quadric states in unobserved views. This, in turn, leads to significant variations in the origin of the object-centric coordinate system in each frame, causing the rigid assumption underlying the object-centric representation of the quadric to no longer hold.

Here, we assume that the object poses obtained through short-term sliding window optimization have been well-optimized. In other words, the poses optimized through the short-term sliding window can serve as priors to guide the subsequent optimization process. This prior constraint can be expressed as:
\begin{equation}\label{16}
\begin{aligned}
e_{\mathcal{C}}\left(_i{sC}^k, _i{\tilde{C}}^k\right) = _i{C}^k - _i{\tilde{C}}^k
\end{aligned},
\end{equation}
where $_i{C}^k, _i{\tilde{C}}^k$ denote the axis length and prior axis of $k^{th}$ object at $i^{th}$ frame respectively.

\section{\bfseries Experiments}
We conduct experiments to evaluate the performance of our proposed method. We test and validate the ego-motion and object motion accuracy of our system using public datasets, simulation datasets and self-built real-world datasets with ground truth annotations of ego-motion and object motion. The experimental operating environment is a laptop computer configured with an Intel Core i7-12700H CPU (14-core 4.7 GHz), 32 GB of RAM, and an NVIDIA GeForce RTX 3060 GPU with 12 GB of graphics memory.
\subsection{\bfseries Evaluation details}
\textbf{Dataset and parameter settings}: For indoor scenes, we evaluate our method using the Oxford Multimotion dataset (OMD) \cite{judd2019oxford}, which is specifically designed for indoor simultaneous camera localization and rigid body motion estimation. The dataset provides ground-truth trajectories obtained through a motion capture system. Furthermore, we have developed synthesized simulation datasets, encompassing both indoor and outdoor scenes, for comprehensive evaluation purposes.

\textbf{Evaluation metrics}: The accuracy of object pose estimation can be evaluated using two different types of metrics: (1) the absolute pose error (APE) and the relative pose error (RPE) measure the quality of the object trajectory; (2) The 3D/2D Intersection over Union (IoU) metric is used to evaluate the accuracy of object scale estimation.

\vspace{-1.6em}
\subsection{\bfseries Oxford Multimotion dataset Evaluation}
\
The proposed method is compared with similar SLAMMOT SLAM systems, including MVO, ClusterVO, and DymSLAM, using the OMD. We follow the evaluation protocol described in \cite{judd2018multimotion}, which involves computing the maximum drift in translation and rotation for both camera ego-motion and all moving objects. Our evaluations and comparisons are conducted on two sequences: swinging 4 unconstrained (SW4) and occlusion 2 unconstrained (O2). The swinging 4 sequence consists of 500 frames with four moving bodies (SW4-C1, SW4-C2, SW4-C3, SW4-C4), while the occlusion 2 sequence comprises 300 frames with two moving bodies (O2-Cuboid and O2-Cube). 

Since we cannot directly obtain the absolute coordinates of the moving objects in the vicon coordinate system, we need to multiply our recovered pose with a rigid transformation matrix $T_{align}$ to align it with the ground-truth pose. This is a reasonable step in the evaluation process because different coordinate systems based on the same trajectory only require a rigid transformation matrix between them.

In the semantic extraction part, we employ the use of point prompt at the keyframe for the AOT. The usage of SAM allows us to perform segmentation on arbitrary objects in the images.

The experimental results is shown as Table.\ref{OMD}, it can be seen that the proposed method achieves more accurate localization performance compared to existing systems in more than 50$\%$ trajectories. 
Furthermore, our approach not only achieves object 3D pose tracking but also facilitates quadric modeling. The modeling results are shown as Fig.\ref{Oxford_SW4} and Fig.\ref{Oxford_O2}. Two main advantages of our method over the existing advanced methods have contributed to this improvement. First, our system optimizes the object states based on the dual sliding window framework, which fully utilizes temporal and spatial continuity constraints. Additionally, compared to existing geometry-based approaches, the robust utilization of semantic information in our system provides better consistency constraints. Second, the powerful segmentation capability of SAM and the pixel-level tracking performance of DeAOT allow our system to effectively suppress the influence of background points in object feature association, thereby avoiding misclassification of dynamic landmarks and degradation of pose estimation results that might occur in geometric-based methods.

Furthermore, the experimental results obtained from the O2 sequence provide further validation for the tracking performance of our meticulously designed AOT. This advanced system enables continuous object tracking even in the presence of temporary occlusions. Illustrated in Fig.\ref{Oxford_O2} subfigures a to d, the swing box undergoes a brief occlusion by the cuboid. However, its motion can still be accurately estimated using the historical observations recorded prior to the occlusion, showcasing the stability of our 3D tracking and scale estimation capabilities.
Moreover, even in scenarios with extensive occlusion, as depicted in subfigure d, where the accuracy of the 2D RotBbox observations diminishes, the estimated scale of the cube remains stable. This observation confirms the advantages of employing quadrics to represent object scales and underscores the robust maintenance of object state consistency achieved through the long-term sliding window approach employed in our system. The more experimental details are shown in the attached video \url{https://youtu.be/_10VeaRSWQ0}.
\vspace{-1.5em}
\subsection{\bfseries Synthesized Multimotion and Modeling dataset Evaluation}
\
Due to OMD lacks ground truth of objects scale, it is not suitable for object modeling accuracy evaluation. 
To overcome the gap, we developed the SMMD (Synthesized Multimotion and Modeling dataset) to extensively evaluate the capability of our method in dealing with 3D traking and modeling of rigid objects with diverse shapes and complex motions. The synthesized dataset includes RGB images, depth images, and simulated LiDAR data, along with ground truth of camera pose, object's pose, and object's scale which can be used to evaluate the ego-localization of SLAM, object 3D tracking, and scale estimation performance. The simulator is released as \url{https://github.com/Linghao-Yang/Synthesized-Multimotion-and-Modeling-Dataset}.

The process of generating simulated LiDAR data is shown in Fig.\ref{LiDARGen}. Firstly, using OpenGL to render a depth image that aligns with the field of view (FOV) of the LiDAR. Subsequently, an octree map is employed to store the point clouds in the camera frame. Finally, employing the predefined LiDAR scan pattern, ray casting is executed to convert the depth image into simulated LiDAR data.
\begin{figure}[t]
\setlength{\abovecaptionskip}{0.cm}
\setlength{\belowcaptionskip}{-0.5cm}
\setcounter{figure}{9}
	\centering
	\includegraphics[width=1.0\linewidth]{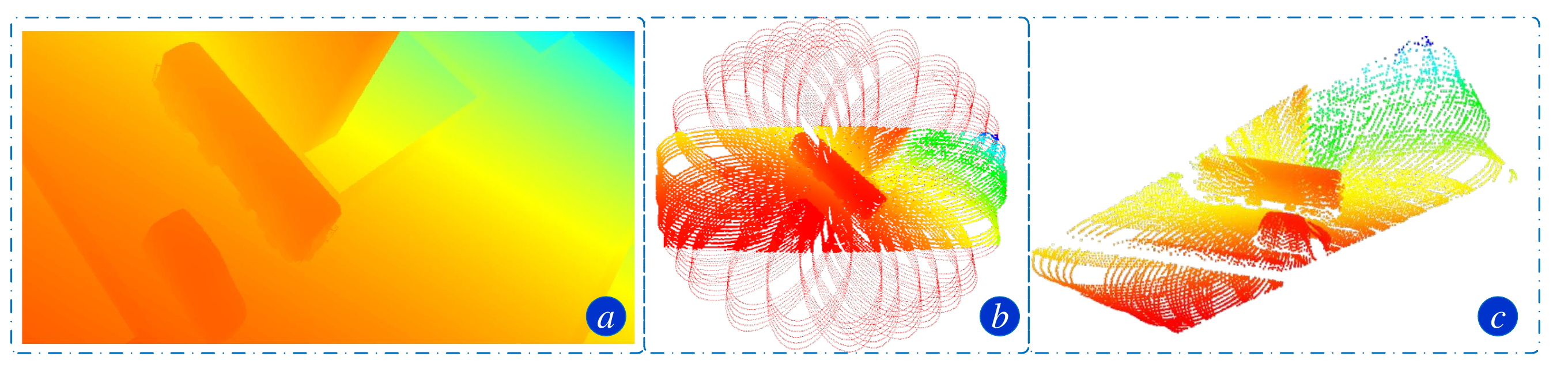}
	\caption{LiDAR data Generation. The first column shows the raw depth image, the second displays ray casting correlation based on the LiDAR pattern, and the third reveals the generated LiDAR point cloud.}
	\label{LiDARGen}
\end{figure}

\subsubsection{\textbf{{Object Pose Tracking Evaluation}}}

The Table.\ref{ObjSynthesizedDataset} quantitatively presents the comparison results of object pose estimation on the synthetic dataset. 
In the first two sequences (Subway $\&$ car), simulated LiDAR is utilized for recovering depth, while in the subsequent sequences, depth images are employed. Regarding depth noise, we assume that the uncertainty of depth values is directly proportional to their distance from the camera's optical center. In other words, the farther the distance from the optical center, the larger the depth noise. 
\vspace{-1em}
\begin{table}[h]\tablefont
\begin{center}
\caption{Performance comparison in synthesized dataset.}\label{ObjSynthesizedDataset}
\setlength{\belowcaptionskip}{-0.5cm}   
\resizebox{\linewidth}{!}{
\begin{tabular}{c|c c c c|c c c c}
\hline
\rowcolor[HTML]{E6E6E6}
\multirow{2}{*}{Sequence} & \multicolumn{4}{c|}{Single Frame OBB} & \multicolumn{4}{c}{Proposed Approach} \\
\rowcolor[HTML]{E6E6E6}
\multirow{-2}{*}{Sequence}
& APE.t & APE.r & RPE.t & RPE.r & APE.t & APE.r & RPE.t & RPE.r\\
\hline
Subway & 4.819 & 2.363 & 6.215 & 1.021 & \underline{\textbf{0.069}} & \underline{\textbf{0.397}} & \underline{\textbf{0.055}} & \underline{\textbf{0.003}}\\
\rowcolor[HTML]{F5F5F5}
car & 2.414 & 2.451 & 3.381 & 0.933 & \underline{\textbf{0.244}} & \underline{\textbf{0.066}} & \underline{\textbf{0.261}} & \underline{\textbf{0.012}}\\
Cuboid & 0.651 & 2.387 & 0.317 & 1.014 & \underline{\textbf{0.082}} & \underline{\textbf{0.144}} & \underline{\textbf{0.034}} & \underline{\textbf{0.018}}\\
\rowcolor[HTML]{F5F5F5}
Cube & 0.44 & 2.26 & 0.411 & 1.51 & \underline{\textbf{0.13}} & \underline{\textbf{0.247}} & \underline{\textbf{0.103}} & \underline{\textbf{0.016}}\\
SkateBoard & 1.111 & 1.218 & 1.173 & 0.299 & \underline{\textbf{0.118}} & \underline{\textbf{0.117}} & \underline{\textbf{0.118}} & \underline{\textbf{0.006}}\\
\rowcolor[HTML]{F5F5F5}
Ball & 0.248 & 2.370 & 1.572 & 1.692 & \underline{\textbf{0.186}} & \underline{\textbf{0.182}} & \underline{\textbf{0.223}} & \underline{\textbf{0.017}}\\
\hline
\end{tabular}
}
\end{center}
\end{table}
\vspace{-1em}
\begin{figure*}[ht]
\setlength{\abovecaptionskip}{0.cm}
\setlength{\belowcaptionskip}{-0.2cm}
\setcounter{figure}{11}
	\centering
	\includegraphics[height = 6.5cm,width=1.0\linewidth]{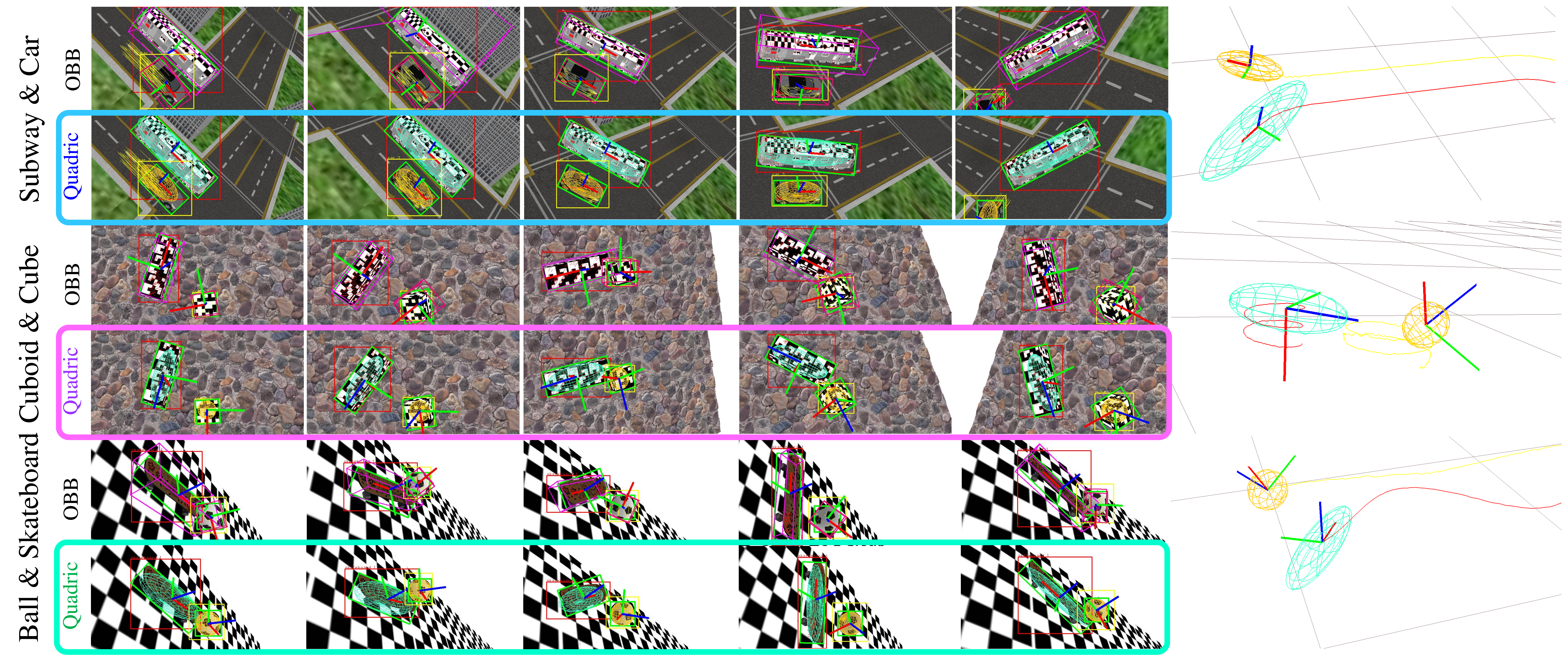}
	\caption{The qualitative results demonstrate the comparison of pose estimation accuracy and modeling accuracy for objects under different algorithms. The rows highlighted with colored boxes in the figure represent the qualitative results of our algorithm.}
	\label{Sim_Map_Eva}
\end{figure*}
\begin{figure*}[t]
\setlength{\abovecaptionskip}{0.cm}
\setlength{\belowcaptionskip}{-0.4cm}
\setcounter{figure}{12}
	\centering
	\includegraphics[height = 6.5cm,width=1.0\linewidth]{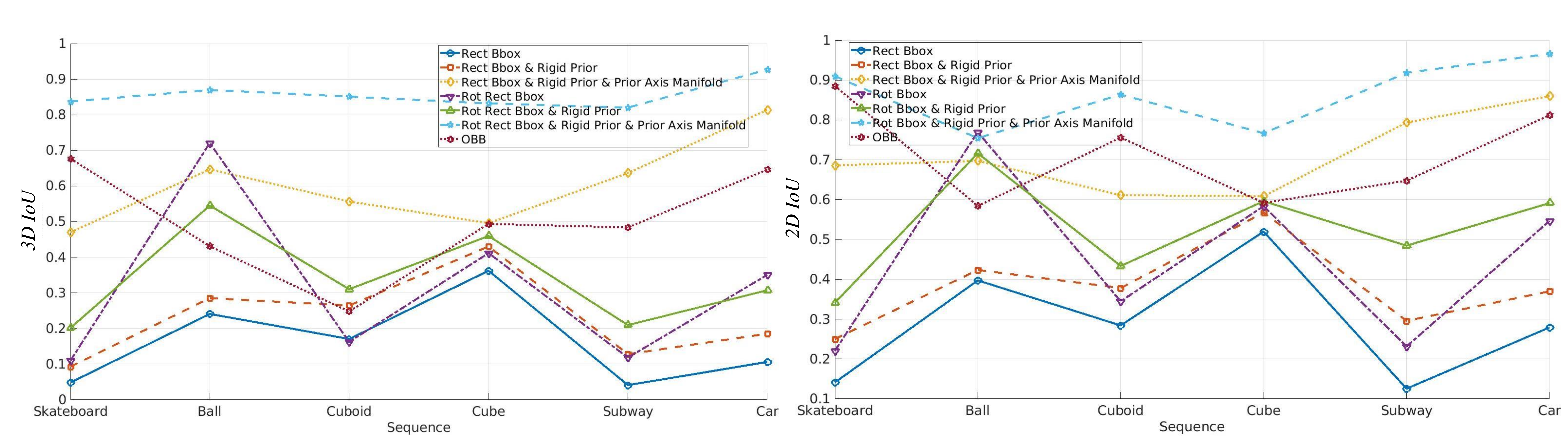}
	\caption{The quantitative results illustrate the comparison of modeling accuracy for objects under different constraints or methods.}
	\label{Sim_IoU_Eva}%
\end{figure*}

We conducted a comparative analysis between our proposed method and the OBB Fit Alg.\ref{OBBRANSAC}. The experimental results clearly demonstrate that our approach achieves a substantial enhancement in object localization accuracy, as indicated by the improvements in APE and RPE compared to solely fitting the OBB. One of the key factors contributing to this improvement is the incorporation of a multi-source observations factor graph model within a dual sliding window framework. The utilization of accurate and continuous observations empowers our system with greater robustness compared to algorithms (e.g., ClusterVO and DynaSLAM2) that rely solely on single-frame, single-source observations. Furthermore, the integration of continuous and stable long-term semantic information for tracking enables better local consistency within our system, which is reflected in the significant improvement of our system in the RPE.

\subsubsection{\textbf{Modeling Accuracy Evaluation}}

Since our method models objects as quadrics, the conventional 3D bounding box IoU evaluation strategy is not appropriate for our method. Consequently, we introduced new evaluation metrics based on 3D IoU and 2D IoU.

Due to the irregular geometry formed by the intersection of two ellipsoids with different coordinate origins, it is challenging to calculate the volume analytically. Therefore, we employ the Monte Carlo algorithm to compute the 3D IoU. In Fig.\ref{SimMetric}, the blue ellipsoid represents the estimated ellipsoid, the red ellipsoid represents the reference ellipsoid, and the green point cloud represents the points within the intersection region.
For the 2D IoU, we project the quadric onto a rotated rectangle and utilize the Vatti clipping algorithm to calculate the intersection area.
\begin{figure}[h]
\setlength{\abovecaptionskip}{0.cm}
\setlength{\belowcaptionskip}{-0.5cm}
\setcounter{figure}{10}
	\centering
	\includegraphics[width=1.0\linewidth]{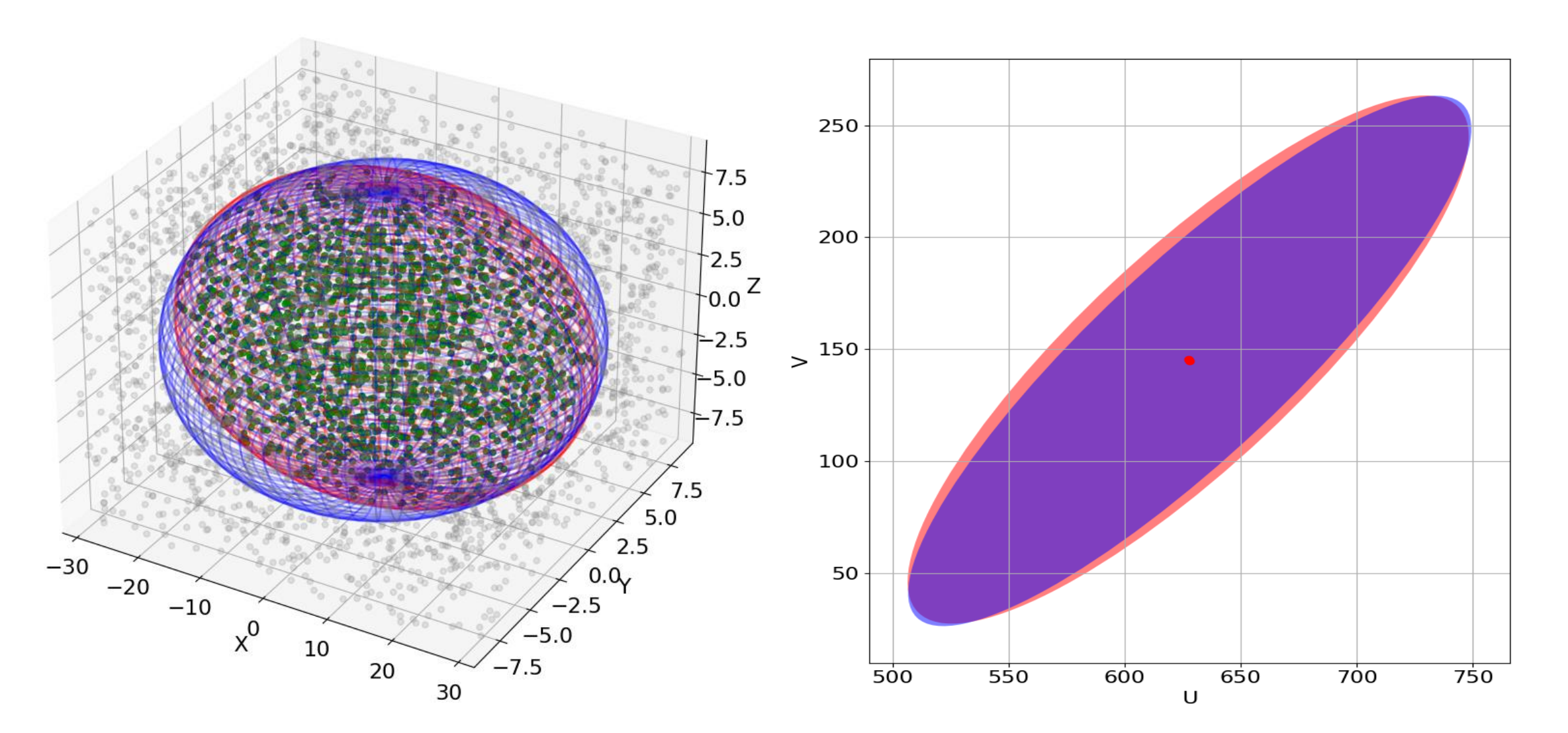}
	\caption{Evaluation Metrics Visulization. The first column represents 3D ellipsoid IoU, the second column represents 2D ellipse IoU.}
	\label{SimMetric}
\end{figure}
\begin{figure*}[ht]
\setlength{\abovecaptionskip}{0.cm}
\setlength{\belowcaptionskip}{-0.5cm}
\setcounter{figure}{13}
	\centering
	\includegraphics[height = 4cm,width=1.0\linewidth]{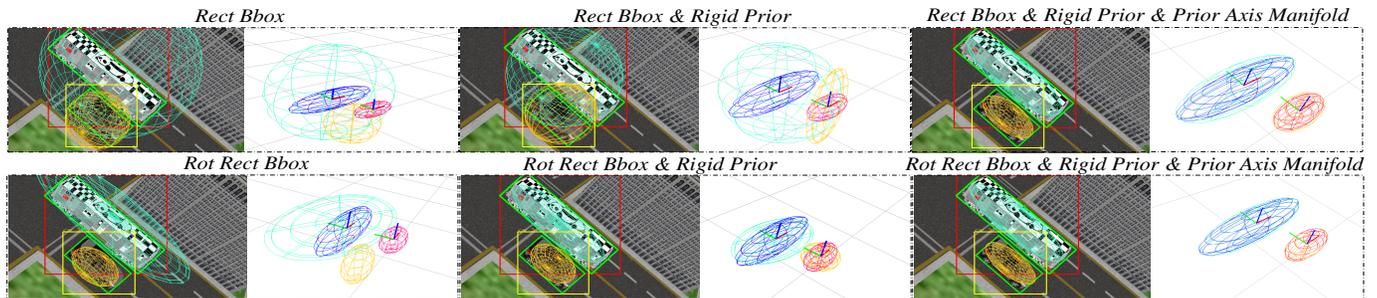}
	\caption{Ablation Study for Scale Constraints. The top row displays results from left to right: combining the rect bbox, rigid prior constraint, and prior axis manifold constraint for scale estimation of the moving object. The bottom row shows results using the rot rect bbox as the observation.}
	\label{Ablation}
\end{figure*}

The Fig.\ref{Sim_Map_Eva} showcases the qualitative IoU comparison results of our system in object pose and scale estimation. It is evident that our algorithm achieves consistent and reliable pose estimation with minimal inter-frame jitter. Furthermore, the fusion of multi-frame observation constraints and the utilization of the compact quadric representation enables robust scale estimation, even in scenarios where tracked objects are partially occluded. The further experimental details are shown in the attached video \url{https://youtu.be/DvanqHV9KNc}.

Additionally, the Fig.\ref{Sim_IoU_Eva} presents a quantitative comparison of the object modeling accuracy on the generated synthetic dataset, we performed ablation experiments to assess the influence of the constraints employed in the scale estimation component. The experimental results demonstrate that incorporating observations based on rotated rectangles, along with the constraints of axis length priors and rigid priors leads to more precise scale estimation of the objects, the average 3D IoU and 2D IoU on all sequences reached over 80$\%$. The constraints of various observations on the scale will be thoroughly analyzed in the upcoming experiments.

\subsubsection{\textbf{Hybrid quadric constrains ablation study}}

\begin{figure}[t]
\setlength{\abovecaptionskip}{0.cm}
\setlength{\belowcaptionskip}{-0.6cm}
\setcounter{figure}{14}
	\centering
	\includegraphics[width=1.0\linewidth]{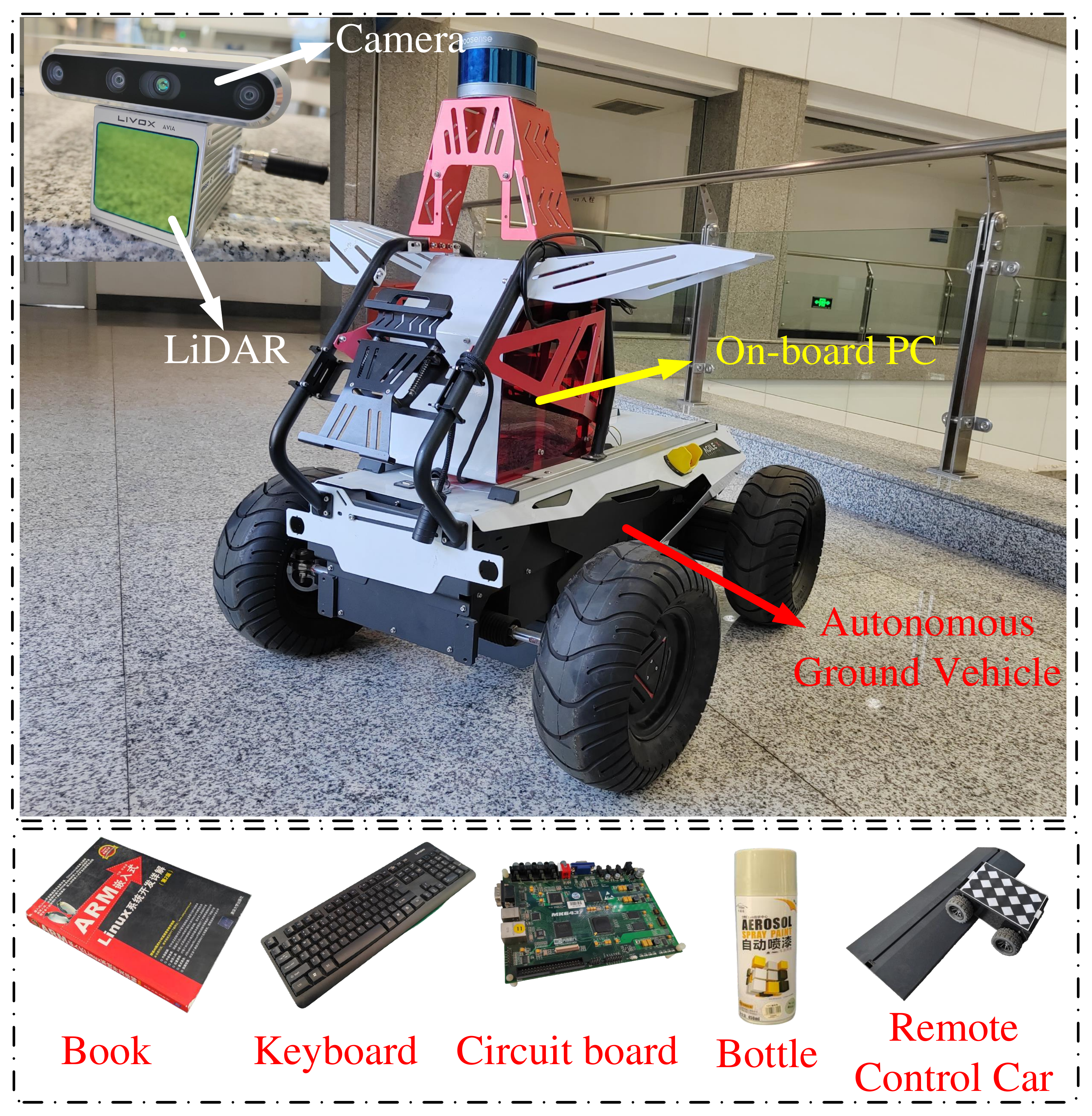}
	\caption{The passive objects used for indoor and outdoor tracking.}
	\label{ExperimentalPlatform}
\end{figure}
\begin{figure}[h]
\setlength{\abovecaptionskip}{0.cm}
\setlength{\belowcaptionskip}{-0.5cm}
\setcounter{figure}{15}
	\centering
	\includegraphics[width=1.0\linewidth]{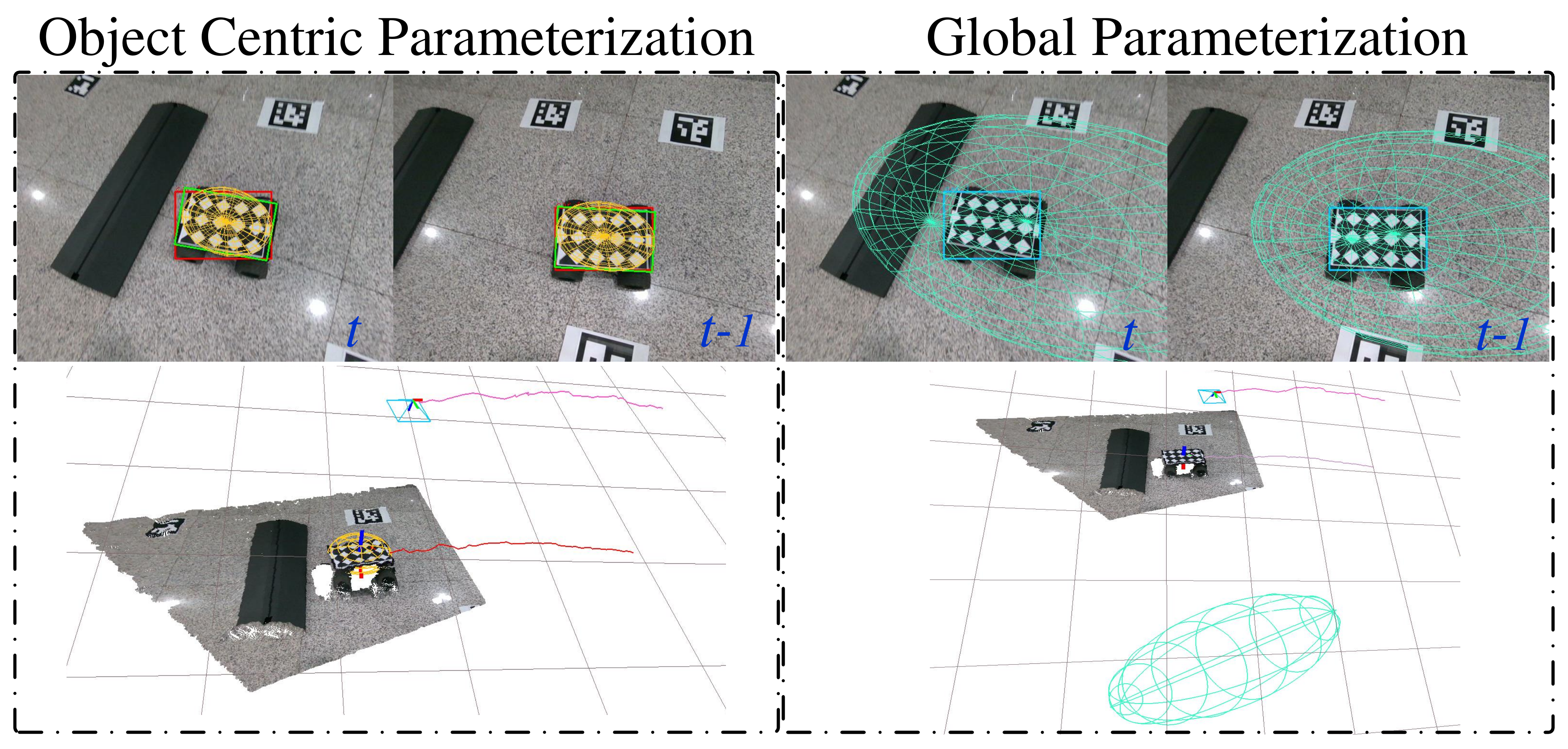}
	\caption{Ablation Study for object-centric parameterization. The first column utilizes the proposed object-centric quadratic parameterization method, while the second column follows global parameterization like other existing approaches.}
	\label{Static_Dyna_Rep}
\end{figure}
The results of the ablation study on hybrid quadric constraints are presented in Fig.\ref{Ablation}. From the findings, it is evident that when utilizing only 2D bounding box information as a constraint, RotBbox observations demonstrate superior constraint effectiveness compared to Bbox, particularly when the ego platform and the observed object are not situated on the same plane and limited viewing angles are available. This superiority can be attributed to the fact that, from the perspective of the limited observation angle, RotBbox observations more accurately represent the actual position of the object on the image plane. Conversely, under conditions of limited viewing angles, Bbox cannot provide sufficient constraints. This limitation arises due to the fundamental principles underlying the multi-view reconstruction process based on quadric, as 2D constraints from restricted views result in under-constrained 3D recovery.
The right column of the figure displays the results of 3D quadric recovery. The blue and red ellipsoids represent the reference ground truth, while the cyan and orange ellipsoids represent the estimation results. It is apparent that although the RotBbox observations appear to align well with the 2D projected quadric curves, the shape becomes under-constrained in 3D space. The recovered ellipsoids significantly deviate from the ground truth in terms of axis lengths, and due to insufficient constraints arising from the limited viewing angles, there is a shift in the object's coordinate origin. This issue arises because, in the current object-centric modeling approach, the axis lengths and poses are interconnected. When the axis length estimation is inaccurate, the object's position is adjusted to achieve the best 2D projection, leading to an incorrect overall result.
\begin{figure*}[t]
\setlength{\abovecaptionskip}{0.cm}
\setlength{\belowcaptionskip}{-0.3cm}
\setcounter{figure}{16}
	\centering
	\includegraphics[width=1.0\linewidth]{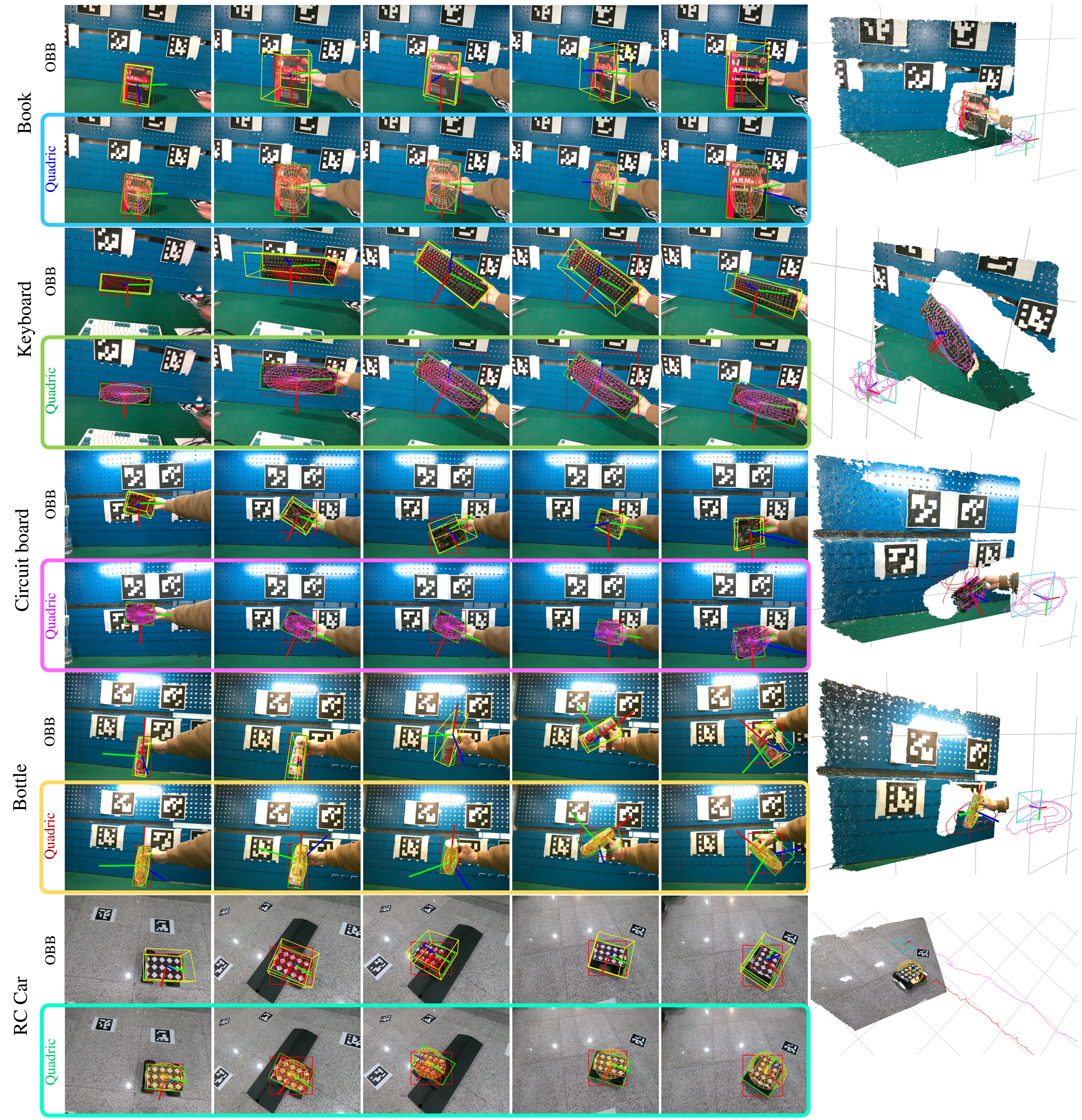}
	\caption{Qualitative evaluation on real-world indoor dataset. The rows highlighted with colored boxes in the figure represent the qualitative results of our algorithm.}
	\label{RGBD_real_world}
\end{figure*}
Just as discussed in Sec.\ref{Rigid_prior_factor},  we introduce the rigid prior assumption to constrain the 3D recovery process under limited viewing angles. 
From the experimental results, it is evident that this approach maintains accurate pose estimation results and improves the estimation of axis lengths. However, it does not entirely solve the under-constrained problem caused by limited viewing angles with only 2D observations.
Therefore, we introduce the 3D prior axis length manifold constraint to tackle this issue. As described in Alg.\ref{MREPAE}, this constraint integrates multiple frame observations and provides precise prior constraints for quadric axis length estimation. The results after incorporating the prior axis length manifold constraint are displayed in the third column of Fig.\ref{Ablation}. It can be observed that with the inclusion of 3D information, the under-constrained problem caused by limited viewing angles is effectively alleviated, resulting in more accurate quadric reconstruction.
\vspace{-1em}
\begin{figure*}[t]
\setlength{\abovecaptionskip}{0.cm}
\setlength{\belowcaptionskip}{-0.5cm}
\setcounter{figure}{18}
	\centering
	\includegraphics[width=1.0\linewidth]{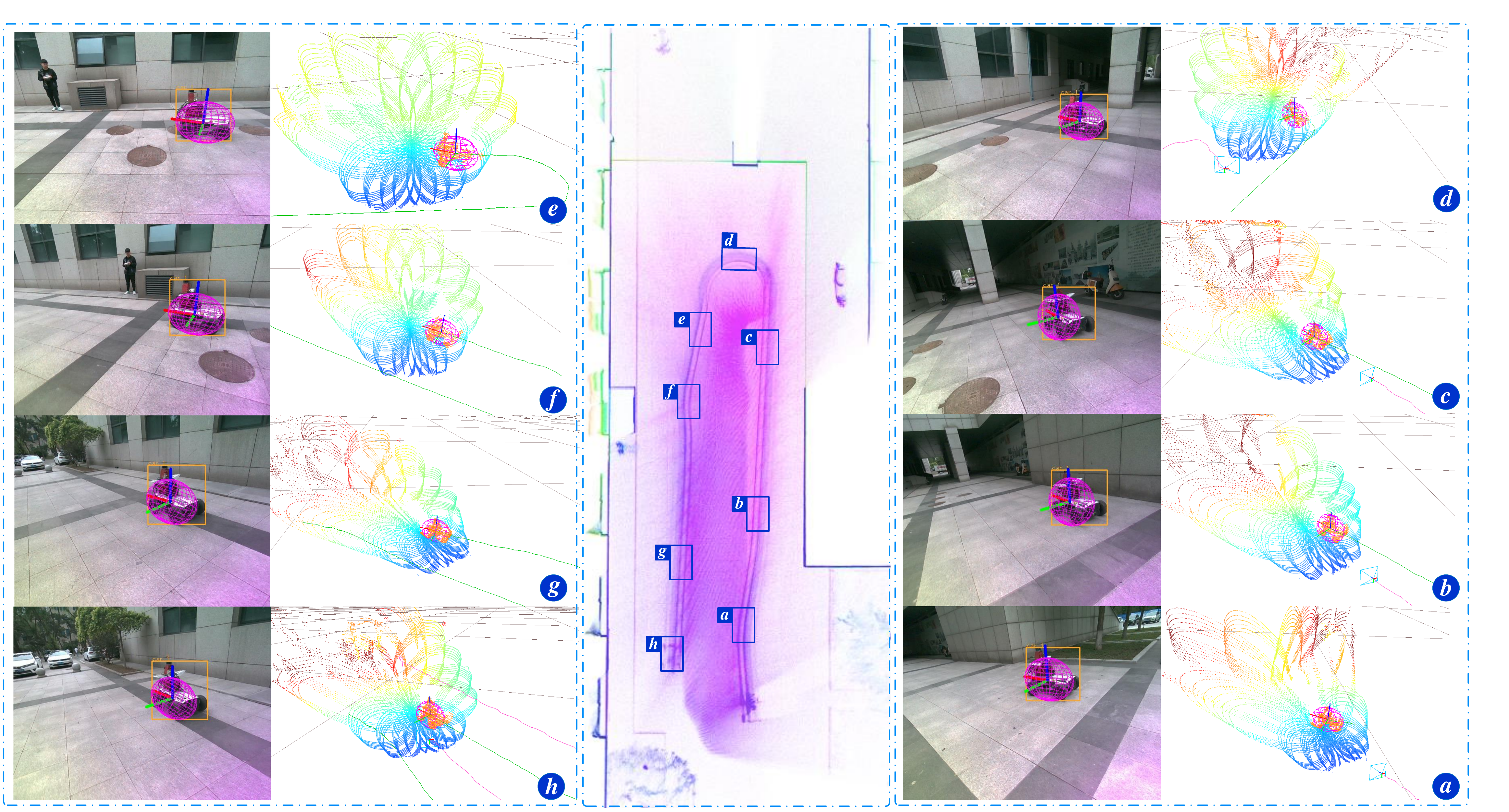}
	\caption{Qualitative evaluation on real-world outdoor dataset.}
	\label{LiDAR_real_world}
\end{figure*}
\subsection{\bfseries Real-world dataset Evaluation}
Furthermore, we conducted verification and evaluation of our method on real-world scenarios. The experimental platform is shown as Fig.\ref{ExperimentalPlatform}, showcasing the input from an indoor scene captured by the RealSense D455 camera, along with the outdoor scene input from the Livox Avia LiDAR and a monocular camera. The objects depicted in the Fig.\ref{ExperimentalPlatform} are used as passive objects for tracking and modeling. 

\subsubsection{\textbf{Object-centric and Global Parameter Comparison}}
Fig. \ref{Static_Dyna_Rep} presents a qualitative comparison between our proposed object-centric quadric parameterization and the original global parameterization approach utilized in existing methods (e.g., \cite{oaslam}) for modeling moving objects. The global parameterization form fixes the global coordinate system begin at the initialization of the quadric, which limits its ability to model moving objects. In contrast, our method allows the quadric to adapt to the motion of objects by representing each object's pose individually, decoupled from the global coordinate system, which makes we can effectively handle both static and moving objects within a unified framework.

\subsubsection{\textbf{Indoor Evaluation}}

The Fig.\ref{RGBD_real_world} qualitatively demonstrates the modeling and pose tracking performance of our system by using an RGB-D camera in indoor scenes. The unified quadric parameterization enables seamless switching between static and dynamic objects, leading to strong consistency in pose estimation with minimal jitter.

In terms of pose estimation, our exhibits robust performance with accurate tracking and low jitter. 
Indeed, solely relying on single-frame observations for OBB fitting is susceptible to the impact of masking observation noise, leading to significant inter-frame jumps in the estimated results. 
Additionally, the formulation of unified pose and scale estimation enables us to address the challenges of estimating the state of both static and dynamic objects simultaneously. Combining the ablation study shown in Fig.\ref{Ablation} and above results, we can conclude that the hybrid constraints contribute to our method's ability to achieve fast scale convergence and accurate object modeling. Moreover, benefit to the dual-sliding window approach in maintaining object states, our system can leverage more observations to enhance the consistency of state estimation and effectively handle occlusion issues.

\subsubsection{\textbf{Object Motion-Aided Undistortion Comparison}}
\begin{figure}[t]
\setlength{\abovecaptionskip}{0.cm}
\setlength{\belowcaptionskip}{-0.5cm}
\setcounter{figure}{17}
	\centering
	\includegraphics[width=1.0\linewidth]{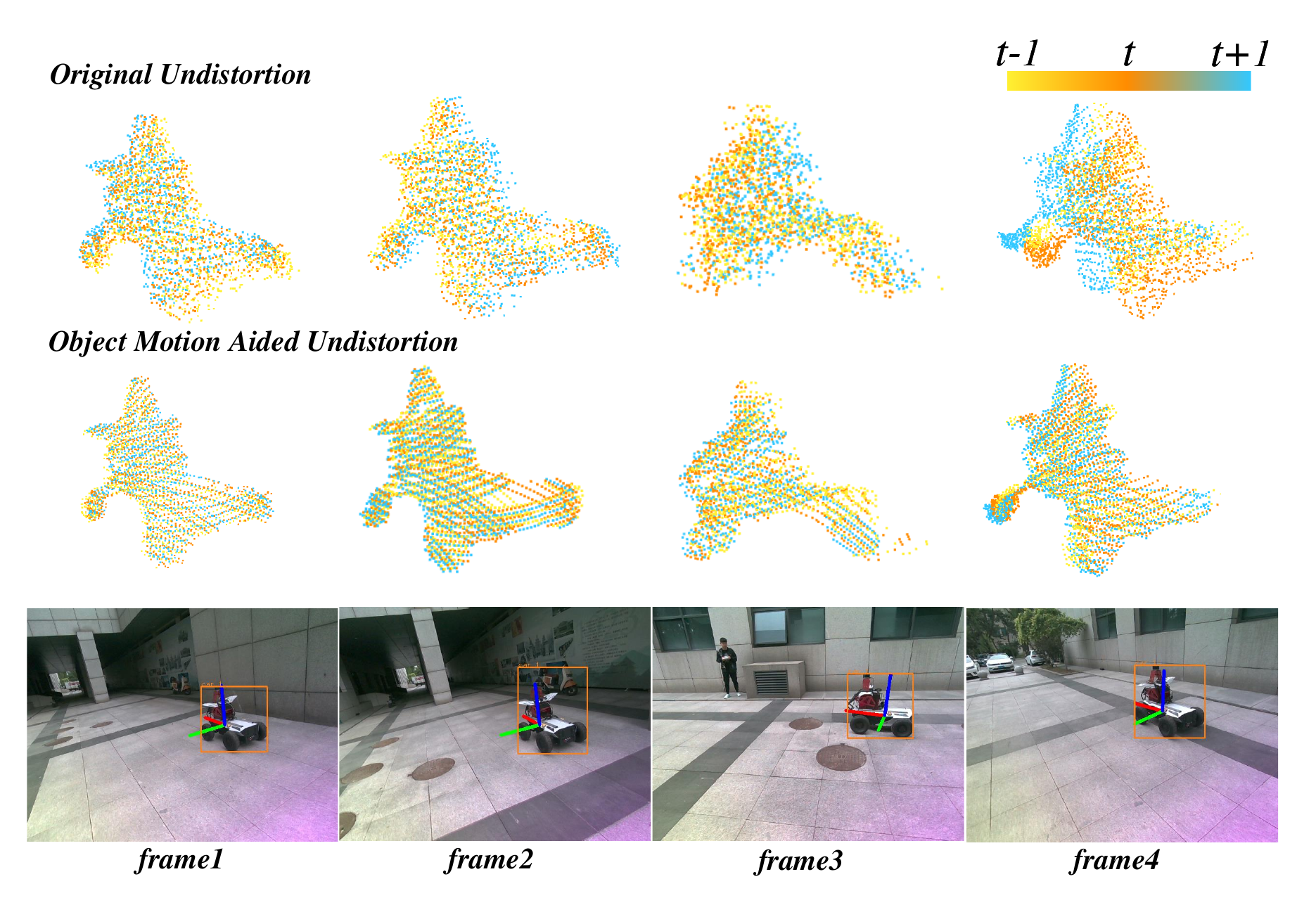}
	\caption{Ablation Study for Object Motion-Aided Undistortion. The original undistortion method's results are shown at the top and the proposed method are displayed at the second row. The bottom row illustrates the corresponding scenarios.}
	\label{OMAU}
\end{figure}

As described in Sec.\ref{outdoor_ego_state_estimation}, to enhance the precision of feature depth estimation in outdoor environments, we have adopted the sensor configuration of Solid LiDAR combined with a monocular camera. The experiments in this subsection are intended to validate the Object Motion-Aided Undistortion algorithm proposed in this paper, and the experimental results are shown in Fig.\ref{OMAU}. The first two rows display the point cloud correction results using both the original undistortion method and our proposed method for three consecutive frames, while the third row depicts the corresponding real-world scene.
The experimental results reveal that slower object motion and a higher resemblance between ego motion and object motion lead to less impact on the undistortion process. However, when these conditions are not met, it significantly impairs undistortion performance, causing inaccurate feature depth estimation and, in turn, affecting subsequent state estimation. By introducing the feedback of the object perception results into the point cloud undistortion process, we are able to decouple the motion and obtain a cleaner and more accurate point cloud. 
\subsubsection{\textbf{Outdoor Evaluation}}
Fig.\ref{LiDAR_real_world} depicts the object perception results obtained from our method in a real-world outdoor scene. The orange point cloud represents the object point cloud generated by our system. Notably, by employing a soft timestamp synchronization and motion compensation algorithm as detailed in Sec.\ref{outdoor_ego_state_estimation}, we significantly reduce distortions in the acquired point clouds. This refinement enhances the reliability of the depth information obtained. Moreover, we successfully leverage the robustness of image information in terms of texture feature association, enabling our system to achieve smooth and accurate 3D object tracking while effectively handling complex real-world scenes. The more experimental details are shown in the attached video \url{https://youtu.be/_b7f3Wzr7nE}.
\vspace{-1em}
\section{\bfseries conclusion}
In this paper, we present a novel optimization framework that integrates the 3D tracking and modeling of both static and dynamic rigid objects. Through a comprehensive analysis, we identify the limitations of the original global parameterization method for quadric in the dynamic object modeling problem. To address these limitations, we introduce an object-centric dual quadric parameterization, which allows for the estimation of both static and dynamic quadric within a unified model. Our framework outperforms existing methods on diverse datasets by leveraging the SQI algorithm and a 9 DoF object state estimation algorithm based on the dual sliding window framework with hybrid constraints. Moreover, we extend the applicability of our framework to indoor and outdoor environments by proposing solutions that combine pure vision and visual-LiDAR fusion.
In future research, our main objectives are to improve the accuracy and precision of pose estimation for objects with low-texture surfaces. Additionally, we aim to integrate the estimated object state into planning and decision-making algorithms, enabling the development of an accurate and robust intelligent tracking and control system.

\bibliographystyle{ieeetr} 
\bibliography{paperlist}

\end{document}